\documentclass[sigconf,nonacm]{acmart}

\AtBeginDocument{%
  }

\usepackage{amsmath,amsfonts,bm}

\def\eqref#1{equation~\ref{#1}}

\def\1{\bm{1}}

\def\vg{{\bm{g}}}

\def\vv{{\bm{v}}}

\DeclareMathAlphabet{\mathsfit}{\encodingdefault}{\sfdefault}{m}{sl}
\SetMathAlphabet{\mathsfit}{bold}{\encodingdefault}{\sfdefault}{bx}{n}

\newcommand{\ie}{i.e.\ }
\newcommand{\eg}{e.g.\ }

\renewcommand{\aa}{\bm{a}}
\newcommand{\FF}{\bm{F}}

\usepackage{subcaption}
\usepackage{siunitx}

\usepackage{calc}

\usepackage{enumitem}

\usepackage{soul}

\makeatletter
\@ACM@authorversiontrue

\makeatother

\begin{document}

\title{Reward Function Design for Crowd Simulation via Reinforcement Learning}


\author{Ariel Kwiatkowski}
\email{ariel.kwiatkowski@polytechnique.edu}
\affiliation{%
  \institution{Institut Polytechnique de Paris}
  \city{Palaiseau}
  \country{France}
}

\author{Vicky Kalogeiton}
\email{vicky.kalogeiton@polytechnique.edu}
\affiliation{%
  \institution{Institut Polytechnique de Paris}
  \city{Palaiseau}
  \country{France}
}

\author{Julien Pettré}
\email{julien.pettre@inria.fr}
\affiliation{%
  \institution{INRIA}
  \city{Rennes}
  \country{France}
}

\author{Marie-Paule Cani}
\email{marie-paule.cani@polytechnique.edu}
\affiliation{%
  \institution{Institut Polytechnique de Paris}
  \city{Palaiseau}
  \country{France}
}


\renewcommand{\shortauthors}{Kwiatkowski et al.}

\begin{abstract}
Crowd simulation is important for video-games design, since it enables to populate virtual worlds with autonomous avatars that navigate in a human-like manner. 
Reinforcement learning has shown great potential in simulating virtual crowds, but the design of the reward function is critical to achieving effective and efficient results. In this work, we explore the design of reward functions for reinforcement learning-based crowd simulation. We provide theoretical insights on the validity of certain reward functions according to their analytical properties, and evaluate them empirically using a range of scenarios, using the energy efficiency as the metric. Our experiments show that directly minimizing the energy usage is a viable strategy as long as it is paired with an appropriately scaled guiding potential, and enable us to study the impact of the different reward components on the behavior of the simulated crowd. Our findings can inform the development of new crowd simulation techniques, and contribute to the wider study of human-like navigation.
\end{abstract}

\begin{teaserfigure}
\centering
\begin{subfigure}{0.19\textwidth}
    \includegraphics[width=\textwidth]{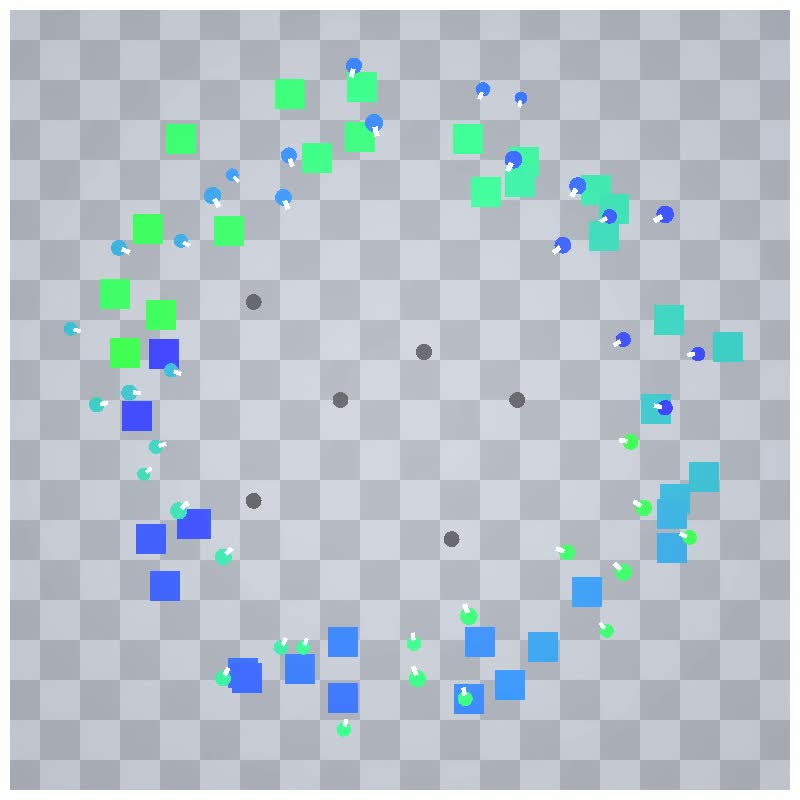}
    \caption{Circle scenario.}
    \label{fig:Circle}
\end{subfigure}
\hfill
\begin{subfigure}{0.19\textwidth}
    \includegraphics[width=\textwidth]{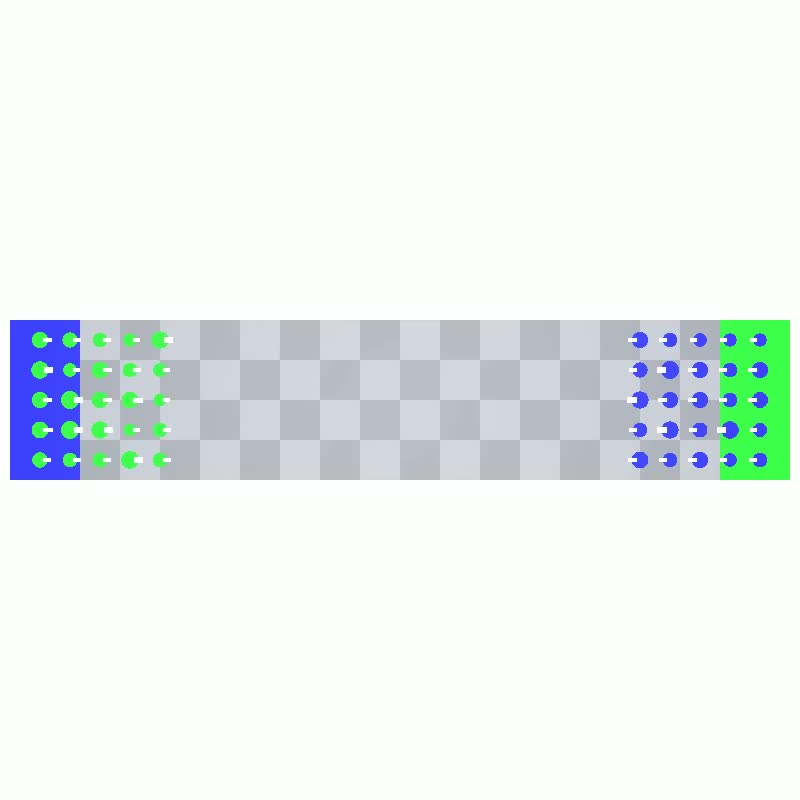}
    \caption{Corridor scenario.}
    \label{fig:Corridor}
\end{subfigure}
\hfill
\begin{subfigure}{0.19\textwidth}
    \includegraphics[width=\textwidth]{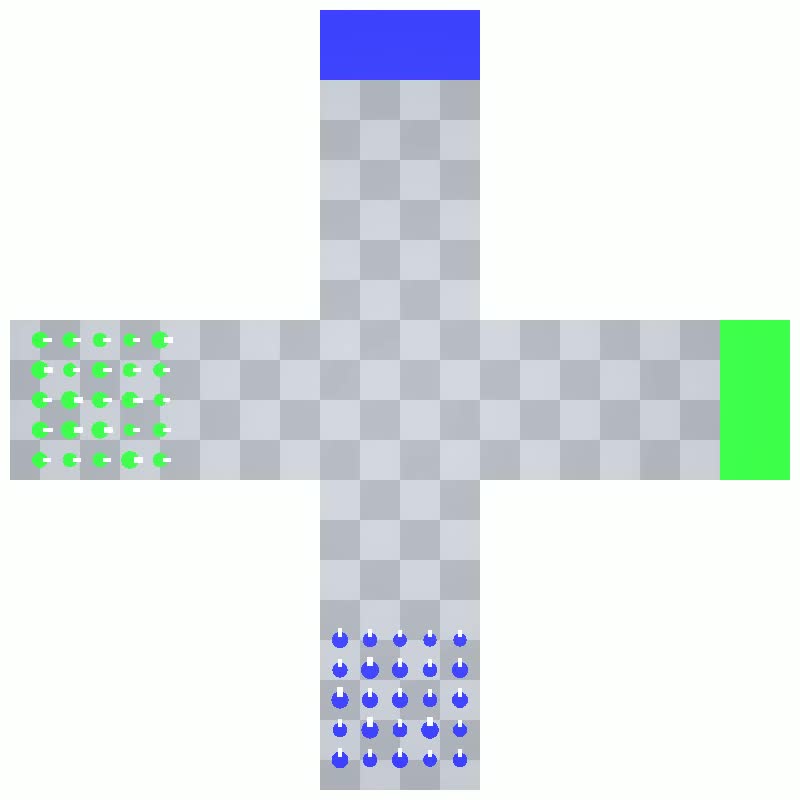}
    \caption{Crossing scenario.}
    \label{fig:Crossing}
\end{subfigure}
\hfill
\begin{subfigure}{0.19\textwidth}
    \includegraphics[width=\textwidth]{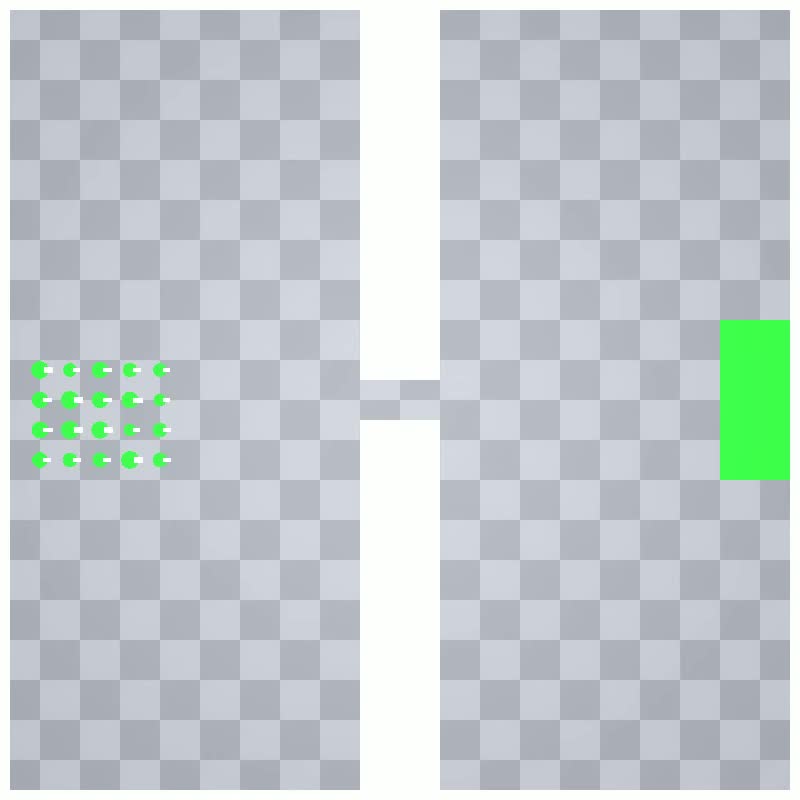}
    \caption{Choke scenario.}
    \label{fig:Choke}
\end{subfigure}
\hfill
\begin{subfigure}{0.19\textwidth}
    \includegraphics[width=\textwidth]{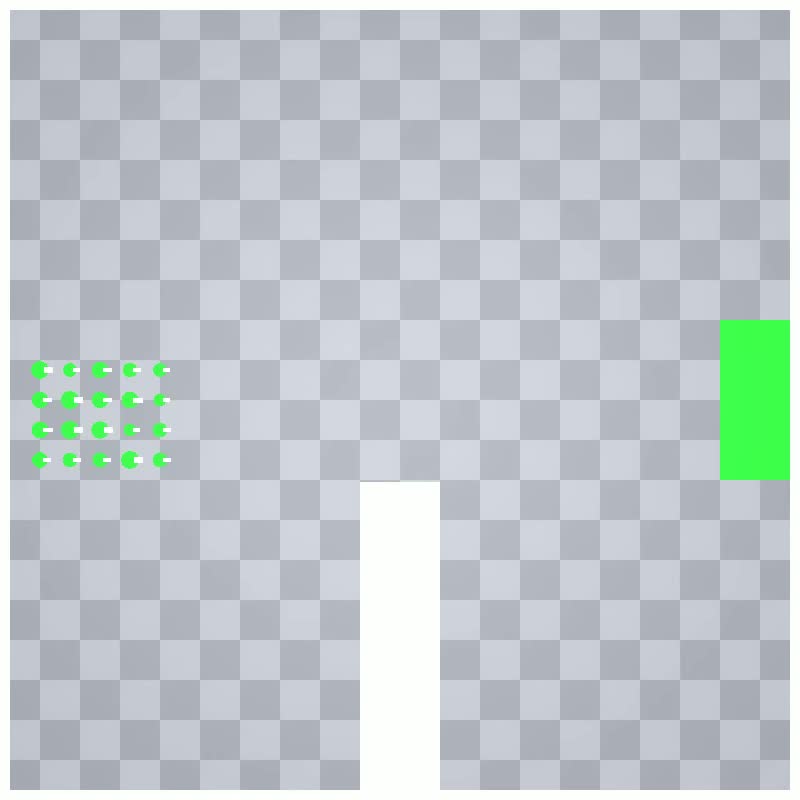}
    \caption{Car scenario.}
    \label{fig:Car}
\end{subfigure}
\caption{
Agent's initial positions and goals in five scenarios:
\textbf{(a)} Circle with 40 agents. \textbf{(b)} Corridor with 50 agents. \textbf{(c)} Crossing with 50 agents. \textbf{(d)} Choke with 20 agents. \textbf{(e)} Car with 20 agents. In each scenario, agents must reach the goal with the same color as them. In the circle scenario, initial starting positions are randomly perturbed during each episode. In the car scenario, the obstacle at the bottom of the scene moves upwards.}
\label{fig:scenarios}
\end{teaserfigure}


\maketitle

\section{Introduction}

Reinforcement Learning (RL) holds a unique potential for simulation of human crowds, offering flexibility and power that traditional control or planning algorithms often lack. However, successfully using RL for this purpose brings about new challenges, primarily rooted in the need to design an effective reward function.

The design of the reward function is crucial for the success of RL algorithms in real-world applications. The balance between sparsity and density of rewards has major implications for the performance of these algorithms. Sparse rewards may lead to the standard algorithms not converging in reasonable time. Conversely, overly dense reward could potentially impact the optimal policy and the relative performances of various suboptimal policies. This issue is particularly relevant in the context of simulating human crowds where, apart from clear objectives like navigation and collision avoidance, the goal of reproducing human-like behavior remains somewhat vague.

During locomotion, humans tend to move at a certain comfortable speed that is specific to the individual, usually around $\SI{1.3}{\meter/\second}$~\citep{whittle_gait_2008}. Following \citet{guy_pledestrians_2010}, this is as a result of minimizing the energy expended when moving between two points. In principle, this measure could be used as a reward function for an RL agent to optimize. In practice, however, this tends to be ineffective due to the unique structure of energy minimization, where agents must take short-term negative rewards to obtain long-term positive rewards. The typical solution is designing an artificial reward function, lacking an explicit connection to the energy minimization aspect, but focusing on rewarding movement towards the goal at the right speed.

We propose the development of a more principled reward function that takes into consideration energy efficiency of motion, serving as a proxy for human-likeness. This choice stems from the lack of metrics that specifically quantify human-likeness in existing literature. 
It is important to note that energy efficiency does not fully describe human behavior, ignoring aspects like long-term goals and subtle inter-personal interactions. Nonetheless, this approach lays the groundwork for more advanced future methods.

We validate our approach both theoretically and empirically. First, we analyze the properties of various reward functions under the discounted utility paradigm. Second, we train RL agents using these reward functions, and compare their performance using the metric of energy usage.

\smallskip
\noindent
Our contributions are:
\vspace{-1.5mm}
\begin{enumerate}
    \item Physically-based extension of the energy usage model that accounts for acceleration.
    \item Evaluation of various reward functions as proxies for energy minimization.
\end{enumerate}


\section{Related Work}

Crowd simulation has gained considerable attention in the field of computer graphics, artificial intelligence, and robotics. Early techniques relied on rule-based systems, and force-based or velocity-based methods (see~\citep{toll_algorithms_2021} for a review). 
Recently, there has been an increasing interest in employing Reinforcement Learning (RL) and Deep Reinforcement Learning (DRL) for crowd simulation~\citep{kwiatkowski_survey_2022}. In this section, we briefly summarize prior work that is relevant to RL crowds simulation.

\textbf{Reinforcement Learning.} RL is an approach to learning sequential decision-making processes, where agents interact with their environment to maximize cumulative rewards. State-of-the-art RL algorithms frequently use neural networks, such as in the Policy Gradient Theorem~\citep{sutton_policy_1999} and Proximal Policy Optimization (PPO) algorithm~\citep{schulman_proximal_2017}. The latter has become the de facto standard on-policy algorithm due to its simplicity and efficiency, and is the algorithm we use in this work.

\textbf{Reward function.} Designing the right reward function is a critical aspect of RL as it shapes the agent's behavior and learning process. It is often nontrivial and requires striking a balance between simplicity and expressiveness~\citep{ng_policy_1999}. Sparse rewards may lead to difficulties in exploration, while overly dense rewards can result in unintended behaviors or suboptimal solutions~\citep{sutton_reinforcement_2018}. Several works have addressed reward function design, including inverse reinforcement learning (IRL)~\citep{ng_algorithms_2000, abbeel_apprenticeship_2004}, which aims to learn the reward function by observing expert demonstrations, and reward shaping~\citep{ng_policy_1999}, which augments the original reward function to guide the agent's learning towards a desired behavior. In multiagent settings, designing the reward function becomes even more challenging, as the interactions between agents need to be considered~\citep{leibo_multi-agent_2017}. As such, the importance of reward function design in RL cannot be understated, as it directly influences the agent's learning efficiency, generalization capability, and ultimately, the quality of the learned policy. In this work, we draw from the idea of using a potential term, and adapt it to the crowd simulation setting.


\textbf{Crowd Simulation via DRL.} Various studies have applied DRL to crowd simulation tasks. \citet{long_towards_2018} focus on multiagent robotic navigation tasks, while \citet{lee_crowd_2018} demonstrate that a single trained RL agent can control multiple agents in diverse crowd scenarios. \citet{sun_crowd_2019} train groups of agents by following leader agents, and other works~\citep{xu_local_2020, zheng_improved_2019} combine DRL with velocity obstacle components for collision-free movement.
To generate high-quality trajectories, \citet{xu_human-inspired_2021} use real-world human trajectory data to train a supervised model that evaluates the human-likeness of generated trajectories. \citet{hu_heterogeneous_2022} and \citet{panayiotou_ccp_2022} employ parametric RL approaches to produce heterogeneous behaviors and configurable agent personalities. \citet{lv_emotional_2022} model realistic crowds in combat simulations using the concept of emotional contagion. \citet{kwiatkowski_understanding_2023} explore the impact of observation spaces on the effectiveness of RL for crowd simulation. In this work, we introduce a more principled approach of designing the reward function for human-like crowds.

\textbf{Energy efficiency.} A commonly used objective for generating and evaluating trajectories is the Principle of Least Effort (PLE). Its origins trace back to \citet{zipf_human_1949}, who proposes that human behavior is broadly characterized by minimizing the perceived effort. Taking energy consumption as a measure of effort, this implies a formulation of human-like trajectories being the energy-efficient ones, which has also been used in prior work on crowd simulation~\citep{guy_pledestrians_2010, bruneau_going_2015}. In our work, we extend this paradigm to also be applicable to training crowds with RL.

\section{Energy Usage Model}\label{sec:energy}


In this work, we follow the hypothesis of the Principle of Minimum Energy (PME) as stated by \citet{guy_pledestrians_2010}, according to which humans tend to choose their trajectories based on minimizing the energy usage. Therefore, we use the energy efficiency as the main benchmark for the quality of a given trajectory. While it does not fully describe human-likeness, it is well-defined and easy to estimate with a simple model.

As a starting point, we consider a model of energy usage based on biomechanical research~\citep{whittle_gait_2008}, and used as a metric in a number of works concerning crowd simulation~\citep{guy_pledestrians_2010, bruneau_going_2015, xu_human-inspired_2021, hu_heterogeneous_2022, kwiatkowski_understanding_2023}. We estimate the energy used in a discrete timestep $\Delta t$ as:
\begin{equation}
    \label{eq:energy_base}
    E = (e_s + e_w v^2) \Delta t
\end{equation}
where $e_s$ and $e_w$ are parameters specific to a given person, with typical values of $e_s = 2.23$ and $e_w = 1.26$ in SI units, computed per unit mass~\citep{whittle_gait_2008}.

It is important to keep in mind that this model does not account for acceleration or turning, and instead only applies to linear motion. In this case, the optimal velocity (\ie one that minimizes the energy usage on a fixed straight trajectory) is $v^* = \sqrt{e_s/e_w}$. This value emerges from integrating the energy usage across the entire path -- moving too quickly uses too much energy, and moving too slowly extends the duration of the trajectory, also increasing the energy usage. 

\subsection{Acceleration correction}\label{sec:energy-acceleration}

\begin{figure}
    \centering
    \includegraphics[width=\linewidth]{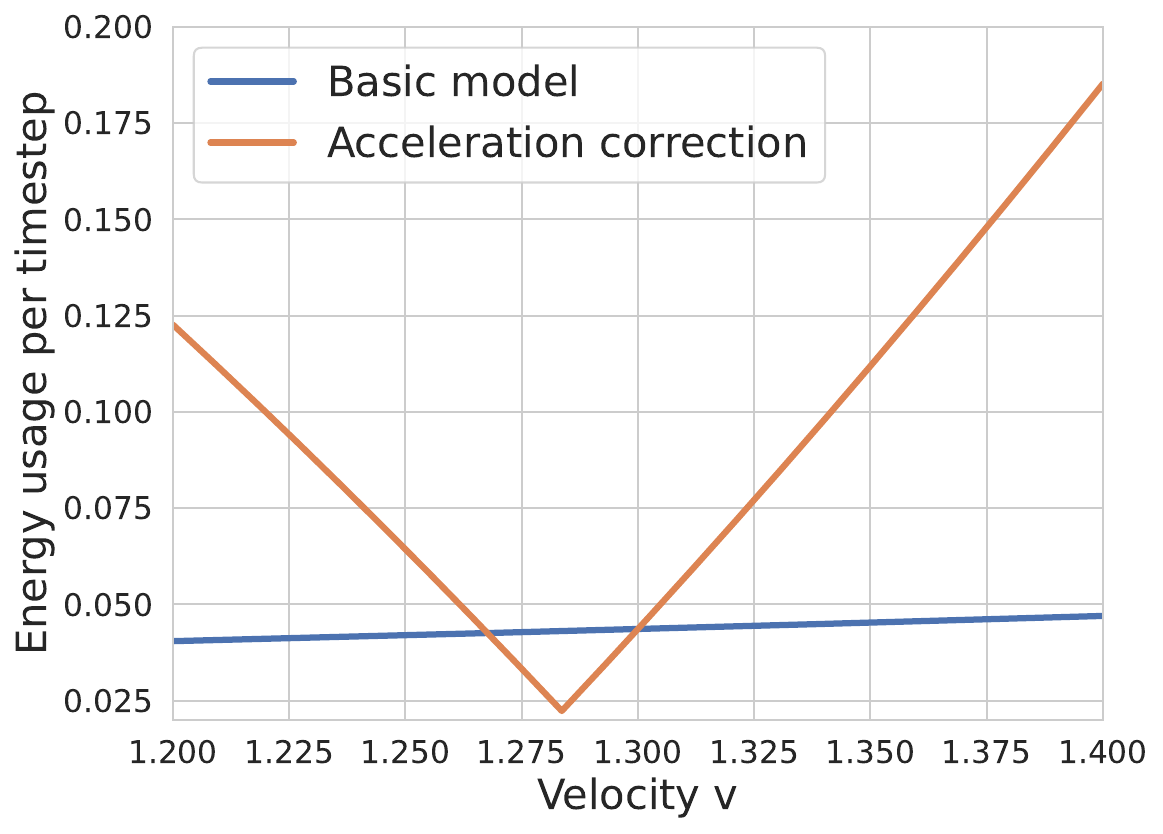}
    \caption{Energy used in a single timestep when moving at a velocity of $v$, after having the velocity of \SI{1.3}{\meter/\second} in the previous timestep, with $\Delta t = \SI{0.01}{\second}$. }
    \label{fig:energy-accel}
\end{figure}

In order to improve the energy estimation, we expand the model in Equation~\ref{eq:energy_base}
so that it also considers the acceleration of agents throughout their trajectories. We start by deriving its basic form. Consider a body moving at a constant velocity $v$, subject to a force opposite to the direction of movement $F_d = -\lambda v$. In Newtonian mechanics, we know that the amount of energy used during displacement is $E = F s$, where $F$ is the applied force, and $s$ is the distance. To adapt this to our discrete model, we factor out the timestep, obtaining $E = F v\ \Delta t$. Substituting the force of drag $F_d$, and setting $\lambda = e_w$ we get:
\begin{equation}
    E = -\lambda v^2 \Delta t = -e_w v^2 \Delta t
\end{equation}

This is the energy lost due to drag in each timestep. To counteract it, the agent needs to use energy equal to the absolute value of this quantity. Combining it with with a constant basal energy usage of $e_s dt$, we get $e = e_s dt + e_w v^2 dt$, recovering Equation~\ref{eq:energy_base}.

To extend this reasoning, consider an agent that moves at velocities $\vv_0$ and $\vv$ in two consecutive timesteps, that is with an acceleration $\aa = \frac{\vv-\vv_0}{dt}$. Assume that the agent is applying a certain force $\FF_a$ in an arbitrary direction in order to modify its velocity. Using simple Euler integration, we have:
\begin{equation}
    \vv = \vv_0 + \FF \Delta t - e_w  \vv_0 \Delta t = (1 - e_w \Delta t) \vv_0 + \FF \Delta t
\end{equation}

Transforming this to obtain the force, we get:
\begin{equation}
    \FF = \frac{1}{\Delta t} (\vv - (1 - e_w \Delta t) \vv_0) = \frac{1}{\Delta t} (\vv - \vv_0 + e_w \vv_0 \Delta t)
\end{equation}

From this we can compute the energy usage as follows:
\begin{align}\label{eq:energy_accel}
    E &= \FF \cdot \vv \Delta t \nonumber\\
    &= \vv \cdot \vv - \vv \cdot \vv_0 + e_w \vv_0 \cdot \vv \Delta t  \nonumber\\
    &= \vv \cdot (\frac{\vv - \vv_0}{\Delta t})\Delta t + e_w \vv_0 \cdot \vv \Delta t \nonumber\\
    &= (\vv \cdot \aa + e_w \vv_0 \cdot \vv) \Delta t
\end{align}

Again taking the absolute value and adding a basal energy usage, we obtain our proposed model for energy usage:
\begin{equation}\label{eq:energy_full}
    E = \left(e_s + |\vv \cdot \aa + e_w \vv_0 \cdot \vv| \right) \Delta t
\end{equation}

\smallskip
To better understand Equation~\ref{eq:energy_full}, 
consider an agent moving with linear acceleration $a$ in the following four cases:
\begin{enumerate} 
    \item Constant motion $a = 0$
    \item Acceleration $a > 0 \iff v > v_0$
    \item Passive deceleration $0 > a > -e_w v_0 \iff v_0 > v > (1 - e_w \Delta t) v_0$
    \item Active deceleration $a < -e_w v_0 \iff v < (1 - e_w \Delta t) v_0$
\end{enumerate}

In the first case $a = 0$, the agent moves at a constant speed $v = ||\vv_0|| = ||\vv_1||$. The energy usage is then:
\begin{equation}
    E = e_s \Delta t + |0 + e_w v^2| \Delta t = (e_s + e_w v^2) \Delta t
\end{equation}
which agrees with Equation~\ref{eq:energy_base}.

If $a > 0$, the agent increases its movement speed. The energy usage then simplifies to:
\begin{align}\label{eq:energy-full}
    E &= (e_s + av + e_w v_0 v)\Delta t\nonumber\\
    &= (e_s + e_w (v - a\Delta t)v + av)\Delta t\nonumber\\
    &= (e_s + e_w v^2 + (1 - e_w \Delta t)av)\Delta t \nonumber\\
    &\approx e_s \Delta t + e_w v^2 \Delta t + av \Delta t 
\end{align}
where the term $av \Delta t$ corresponds to the additional kinetic energy needed to move at a velocity $v$.

If $a < 0$, the agent decelerates. Note, however, that there are 
two distinct possibilities.
If the agent simply stops putting in effort,
it will automatically slow down by a factor of $(1 - e_w \Delta t)$. We call any deceleration below this threshold \textbf{passive deceleration}, which decreases the energy usage. In contrast, if the agent wants to slow down to a speed lower than $(1 - e_w \Delta t) v_0$, this is \textbf{active deceleration}, which requires using additional energy.

We depict this relationship in Figure~\ref{fig:energy-accel}. When the velocity remains constant at $v=v_0=\SI{1.3}{\meter/\second}$, the energy usage is the same in both models. The lowest energy usage (\ie only from the basal metabolic rate) occurs at $v = (1 - e_w \Delta t) v_0 = \SI{1.28}{\meter/\second}$, when the agent decelerates naturally.

\section{Navigation reward design}

Our main goal in this work is designing a reward function which, when optimized, leads to a policy that minimizes the energy usage, as estimated using the model from Section~\ref{sec:energy}. In this section, we discuss a few issues in designing such a reward function.

\subsection{Energy as reward}\label{sec:energy-reward}

A natural starting point is simply using a reward equal to the negative energy usage:
\begin{equation}\label{eq:reward-en-base}
    R = -e_s \Delta t - e_w v^2 \Delta t
\end{equation}
or 
\begin{equation}\label{eq:reward-en-accel}
    R = - e_s \Delta t - |\vv \cdot \aa + e_w \vv_0 \cdot \vv| \Delta t
\end{equation}

This formulation has two critical issues, which make it unfit for being used as a reward function directly. To see this, consider the base reward of Equation~\ref{eq:reward-en-base} for simplicity.

\subsubsection{Local optimum}

In an RL training procedure, each agent begins by taking random actions. In the case of microscopic crowd simulation, that corresponds to choosing a direction, and setting either the velocity or the acceleration in that direction. If an action leads to a higher reward, its probability increases, and if it leads to a lower reward, its probability decreases.

Consider an agent with a simple objective of moving to a specific location, maximizing the reward from Equation~\ref{eq:reward-en-base}. The reward is accumulated from the beginning of the episode, until the agent reaches the goal, or until a predefined time limit. Note that with this structure, the real penalty for not reaching the goal is delivered by the agent having to accumulate the negative reward until the time limit. If the time limit is sufficiently high, it is better for the agent to spend some energy in order to reach its goal and not use any energy afterwards, as compared to spending a long time at rest, even without using energy for movement. 


However, during training, 
the agent is more likely to try to move, but fail reaching the goal.
It then gets the full time-based penalty, but also a penalty for using additional energy for movement. The agent does not know how to reduce the time-based penalty, but it can decrease its energy usage by slowing down. Eventually, it will settle into a local optimum of standing still, which is a failure case.

\subsubsection{Global optimum}

\begin{figure}
    \centering
    \includegraphics[width=\linewidth]{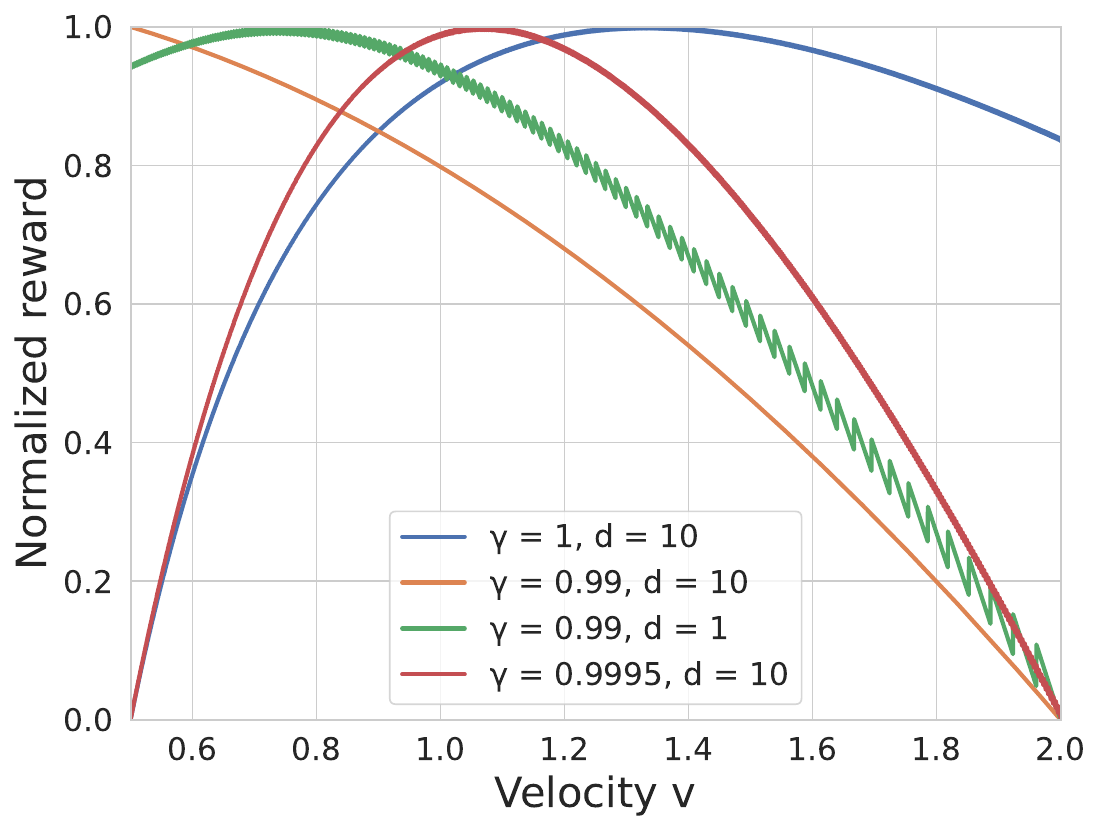}
    \caption{Normalized discounted reward, with energy optimization as the direct objective. Depending on the distance $d$ and the discount factor $\gamma$, the global optimum is different, and in some cases, the optimal behavior is standing still with $v = 0$.}
    \label{fig:energy_global}
\end{figure}

The second problem is related to the fact that modern RL algorithms predominantly use the discounted utility paradigm, weighing future rewards with an exponentially decaying discount factor. Similarly to not reaching the goal, the penalty for moving too slowly is that the agent will have to spend energy in many more timesteps towards the end of the episode. When making a decision at the beginning of the episode, those rewards are heavily discounted, and thus less important.

Consider now the following experiment: the agent travels in a straight line, and has to reach $x = d$ while moving at a constant velocity $v$. The reward is discounted exponentially with a discount factor $\gamma$. In Figure~\ref{fig:energy_global}, we show the discounted reward for some values of $d$ and $\gamma$. The global optimum for a typical discount factor around 0.99 is $v = 0$, which corresponds to the agent not moving at all. Depending on the exact values, the optimal value may be anywhere between $0$ and $\sqrt{\frac{e_s}{e_w}}$, which is a significant problem if our goal is training an agent whose optimal velocity is exactly $\sqrt{\frac{e_s}{e_w}}$.

The fact that discounting changes the optimal policy is not necessarily unexpected. \citet{naik_discounted_2019} show that using discounted rewards when training RL agents may change the optimal policy. In many practical problems, this is not a big concern, and the discount factor is treated as yet another hyperparameter. In this case, however, the discount factor directly impacts the properties of the environment.

\subsubsection{Possible solutions}

There are various ways to tackle the problems described above, but it is important to note that both of them have to be solved together. In order to avoid the local optimum of standing still, we could employ a curriculum-based approach, where the agents initially learn to navigate a short distance without any obstacle. As the training progresses, the distance and the number of agents can be increased, with the hope that the agents will not stop moving.

To fix the issue with the global optimum, the obvious solution is not using any reward discounting. In practice, however, this turns out to be much more unstable and difficult to train. Alternatively, a different non-exponential discounting method could be employed, so that the variance of the gradient estimation is low enough for efficient training, but the optimal velocity remains correct.

Both of these solutions add a non-negligible amount of complexity to the learning algorithm. While in certain situations that might be acceptable, note that all these issues stem from the simple scenario of a single agent navigating to a goal in an energy-efficient manner. With more complicated applications, the complexity is likely to become even higher, \eg via a curriculum designed for a different objective.

To avoid the compounding complexity, we instead propose changing the reward function. Ideally, it should remain similar to the energy usage so that the emergent behavior is still energy-efficient. It should also tackle both of the aforementioned issues -- that is, the reward for moving towards the goal should be higher than for standing still, and the optimal velocity should be invariant under temporal discounting.

\subsection{Energy-based potential}

Adding a guiding potential to the reward function is a common technique of making sparse rewards more dense. \citet{ng_policy_1999} show that adding a reward of the form $R(s, a, s') = \gamma \Phi(s') - \Phi(s)$ does not change the optimal policy for the $\gamma$-discounted rewards. Note that this assumes that the discounted reward is the true objective of the RL task. This is not true in the case of navigation, as we generally want the global energy usage to be optimal. Nevertheless, it can serve as inspiration for designing an analogous guiding term.

In the context of human navigation, there is a simple heuristic that we can use as a guiding potential -- the distance from the goal. Consider the following reward function:

\begin{equation}\label{eq:reward-energy-pot}
    r(\vv) = -e_s \Delta t - e_w v^2 \Delta t + \tilde{c}_p \vv \cdot \hat{\vg}
\end{equation}
where $\hat{\vg}$ is a unit vector pointing from the agent to the goal. Note that the potential term $\vv \cdot \hat{\vg}$ is equal to the change in the distance between the agent and its goal in two consecutive timesteps. 

This induces a total discounted reward of:

\begin{equation}
    R^\gamma = \int_0^T dt( e^{t \ln \gamma} (-e_s \Delta t - e_w v^2 \Delta t + \tilde{c}_p \vv \cdot \hat{\vg}))
\end{equation}

To obtain a bound on the value of $c_p$, we set the condition that when moving directly towards the goal, $R(v^*) > R(0)$, \ie it is better to move towards the goal than stand still. This implies that $\tilde{c}
_p > \sqrt{e_s e_w}$. For simplicity of further analysis, we define $c_p = \frac{\tilde{c}_p}{\sqrt{e_s e_w}}$.

\subsection{Discounting invariance}\label{sec:disc-invariance}

With a simple simulation, it is clear that there is a nontrivial interaction between the values of the discount factor $\gamma$, the coefficient $c_p$, and the optimal velocity $v^\gamma$.

Consider the discounted sum of rewards defined in Equation~\ref{eq:reward-energy-pot}, with a simple policy of moving towards the goal with a speed $v$. With a continuous model of the problem, we can define the discounted sum of rewards as:

\begin{align}\label{eq:discounted-reward-sum}
    R^\gamma &= \int_0^T e^{t \ln \gamma} \left(-e_s - e_w v^2 + c_p \sqrt{e_s e_w} v\right) dt \nonumber \\
    &= \frac{1 - \gamma^\frac{d}{v}}{-\ln \gamma} \left( -e_w v^2 + c_p \sqrt{e_s e_w} - e_s\right)
\end{align}

We differentiate this expression w.r.t. $v$ to obtain an expression for the optimal velocity, and interpret it as an implicit function whose roots correspond to the optimal velocity with a given discount factor $\gamma$:

\begin{equation}
    F(v, \gamma) = \frac{\left( -2
    e_w v + \tilde{c}_p\right) \left(1 - \gamma^\frac{d}{v}\right)}{- \ln \gamma} - \frac{\left( -e_w v^2 + \tilde{c}_p v - e_s \right) \gamma^\frac{d}{v}d}{v^2} = 0
\end{equation}

Solving this analytically for $v$ is difficult. Instead, we consider the implicit derivative:

\begin{equation}
    \frac{dv}{d\gamma} = -\frac{dF}{d\gamma} / \frac{dF}{dv}
\end{equation}

While the resulting expression is highly complex, it is solvable for $c_p$ analytically, yielding the result:

\begin{equation}
    \frac{dv}{d\gamma} = 0 \iff c_p = 2
\end{equation}

This means that using the reward from Equation~\ref{eq:reward-energy-pot} with $\tilde{c}_p = 2\sqrt{e_s e_w}$, the optimal velocity is independent of the discount factor. Note that if we consider non-exponential discounting as a weighted sum of exponential discountings, this conclusion extends to other discounting methods, enabling the application of methods like hyperbolic discounting~\citep{fedus_hyperbolic_2019} or arbitrary non-exponential discounting~\citep{kwiatkowski_ugae_2023}.









\subsection{Non-finishing penalty}\label{sec:non-finish}

When measuring the energy usage as a reward function, or even as a metric, there is another consideration that stems from the RL setting -- the time limit. While theoretically an agent could infinitely explore until they reach the goal, this is impractical. Instead, RL algorithms typically set a maximum number of timesteps allowed in an episode. After this limit passes, the episode terminates, regardless of the state that the agent is in.

In principle, the value of the time limit should not matter as long as it is sufficient to reach the goal. However, the structure of the energy-based reward (Equations~\ref{eq:reward-en-base} and \ref{eq:reward-en-accel}) makes it potentially impactful. Let $T$ be the time limit in seconds, $d$ the total distance from the goal. Moving in a straight line at the optimal velocity $v^*$, the time needed to reach the goal is $T^* = \frac{d}{v^*} = \sqrt{\frac{e_w}{e_s}} d$, and the energy used in this process is $2 \sqrt{e_s e_w} d$. If this energy is greater than that of standing still until the end of the episode $e_s T$, then the optimal policy according to the metric may indeed be simply standing still. 

To prevent this, one option is simply setting the time limit so that $T > 2 \frac{d}{v^*}$, in which case moving at the optimal velocity will result in a lower energy usage than standing still until the end of the episode. This corresponds to an episode length more than twice as long as it would take the agent to reach the goal moving at optimal velocity. A significant drawback of this approach is its inefficiency, as the duration of each episode is significantly extended, which increases the amount of time necessary to collect experience for training. Furthermore, complex scenarios with many agents may extend the optimal trajectories in ways that are difficult to predict before training the agents.

Instead we propose two variants of a heuristic that is added as an additional penalty at the end of the episode if a given agent has not reached its goal. In the first variant, we use the \textbf{optimal} heuristic -- if the agent is at a distance $d$ from its goal, it incurs a penalty of $2\sqrt{e_s e_w} d$, which corresponds to the energy cost it would take to reach the goal moving at the optimal speed in a straight line. In the second variant, instead of using the optimal speed, we use the \textbf{average} speed towards the goal across the agent's trajectory to estimate the remaining energy cost.

Both of these variants have their flaws. Using the optimal heuristic, in certain cases it may be beneficial for agents to only move part of the way, and then stop when they encounter a more dense situation, which requires more energy to navigate. While the average heuristic avoids this issue by directly tying the final penalty to the agent's past performance, the estimated velocity has to be capped at a minimum value (in our experiments: $\SI{0.1}{\meter/\second})$. This avoids issues where the agent has made very little progress towards the goal (which leads to very high penalties, destabilizing the training), or even made negative progress by moving farther from the goal, leading to a negative energy cost and a positive final reward. 

\subsection{Alternative approaches}

In existing literature, most approaches to crowd simulation via RL disregard the problems of energy efficiency, and of encouraging agents to prefer
an intermediate velocity throughout their motion. 
The most common approach to obtain motion with an given velocity $v^*$ is simply setting $v^*$ as the maximum in the environment dynamics~\citep{long_towards_2018, sun_crowd_2019, xu_local_2020, hu_heterogeneous_2022}. 
This is then combined with a guiding potential and a one-time reward for reaching the goal, and due to the incentive structure of the discounted utility paradigm common in RL, this leads to the agents mostly moving at the ``optimal'' (\ie maximum) speed.
The downside of this approach is that agents are unable to move faster than that predefined limit, in contrast with humans, who tend to easily walk, when needed, a little slower or faster than their optimal comfortable speed.

Other works~\citep{lee_crowd_2018, xu_human-inspired_2021, kwiatkowski_understanding_2023} include a velocity-dependent reward term that incentivizes moving at a specific speed which is below the highest allowed speed. Here we analyze and compare each of those approaches.

\citet{lee_crowd_2018} use a function they call FLOOD, defined as follows:
\begin{align}
    &FLOOD(v, v_{min}, v_{mav}) = \nonumber\\ 
    =&|\min(v - v_{min}, 0)|  + |\max(v - v_{max}, 0)|
\end{align}
where $v_{min}, v_{max}$ define the range of comfortable speed. When applied to the linear velocity, this term disincentivizes velocities outside of the preferred range. While this structure is not directly connected with energy optimization, it serves a similar purpose of controlling the movement speed.

A similar structure was used by~\citet{xu_human-inspired_2021}. In their reward function, they use a velocity regularization term:
\begin{equation}
    r(\vv) = \exp\left(\sigma_v ||\vv - \vv^* || \right)
\end{equation}
where $\sigma_v$ is a parameter, and $\vv^*$ is a vector pointing towards the goal, whose magnitude is equal to the optimal velocity. In our energy optimization framework, it is $\vv^* = v^* \hat{\vg} = \sqrt{e_s / e_w} \hat{\vg}$.

Consider the term within the exponent $||\vv - \vv^*|| = ||\vv - v^* \hat{\vg}||$, and take its square. Interpreting this as a scalar product, we have $(\vv - v^* \hat{\vg}) \cdot (\vv - v^* \hat{\vg})$. When we multiply the terms, substitute $v^* = \sqrt{e_s / e_w}$ and use the fact that $||\hat{\vg}|| = 1$, we get $v^2 - 2\sqrt{e_s/e_w} \vv \cdot \hat{\vg} + \frac{e_s}{e_w} = \frac{1}{e_w} \left(  e_s + e_w v^2 - 2\sqrt{e_s e_w} \vv \cdot \hat{\vg} \right)$. This happens to be proportional to the discounting-invariant energy usage with potential. Note, however, that the final reward used by \citet{xu_human-inspired_2021} applies additional operations to this value (square root and exponent).

\citet{kwiatkowski_understanding_2023} use an explicit potential term, and a speed similarity term $c_v |v - v^*|^{c_e}$ which does not take into account the direction of the movement. With the exponent $c_e = 2$, this expands to $\tilde{c}_p \vv \cdot \hat{\vg} - v^2 + 2\sqrt{\frac{e_s}{e_w}} v - \frac{e_s}{e_w}$, which is equal to the energy usage with potential, but with an additional positive term proportional to the agent's speed. This results in a bicycle-like behavior where an agent prefers to artificially extend its trajectory while maintaining its optimal speed, instead of simply slowing down.

\section{Reward evaluation}

In this section, we empirically evaluate our proposed reward structure, and compare it to previously proposed formulations. We also perform an ablation on various parts of the reward function to investigate their importance and impact on the final results.

\subsection{Experimental setup}

We performed the experimental evaluation on five crowd scenarios:
\begin{enumerate}
    \item Circle -- agents start at the perimeter of a circle, and must reach the antipodal point of the circle. We apply noise to both the start and goal positions, and add stationary obstacles in the middle of the circle.
    \item Corridor -- agents start at two ends of a corridor and must reach the opposite end.
    \item Crossing -- agents start at southern and western ends of perpendicularly crossed corridors, and must reach the northern and eastern ends, respectively.
    \item Choke -- agents must pass from west to east through a narrow opening in a wall.
    \item Car -- agents must wait for a moving obstacle to open a passage to the goal.
\end{enumerate}

In each scenario, all agents are given a time limit of 200 time-steps, each lasting \SI{0.1}{\second}. Each agent is removed from the simulation once it touches its goal. Following the classification by \citet{kwiatkowski_understanding_2023}, we use Egocentric observations with Polar Acceleration dynamics. Each agent has randomly sampled parameters of $e_s, e_w$ as defined in Section~\ref{sec:energy}. These values are included in the observation, and used to compute the individual reward of each agent.

The main metric we use for evaluation is Energy+, defined as energy usage with the acceleration correction (Equation~\ref{eq:energy-full}), plus the non-finishing penalty using the average heuristic (Section~\ref{sec:non-finish}). The penalty is meant to additionally penalize agents which do not reach their goals in time, to ensure that agents cannot hack the reward function by stopping in the middle of the trajectory.

\subsection{Reward function structure}

Throughout the various reward functions we evaluate in this work, we use the following components:

\begin{enumerate}
    \item Basal energy usage $r_b = -e_s$  
    \item Velocity-based energy usage $r_v = -e_w v^2$  
    \item Dynamics-based energy usage $r_d = -|\vv \cdot \aa + e_w \vv_0 \cdot \vv|$  
    \item Guiding potential $r_p = 2\sqrt{e_s e_w} \vv \cdot \hat{\vg}$  
    \item Preferred speed matching $r_s = |v - v^*|^{c_e}$  
    \item Speeding penalty $r_z = \max(v - v^*, 0)^{c_e}$  
    \item Exponential velocity matching $r_m = \exp(\sigma_v ||\vv - \vv^*||)$
    \item Final non-finishing penalty using the optimal speed heuristic (Section~\ref{sec:non-finish}) $r_{o}$ 
    \item Final non-finishing penalty using the average speed heuristic (Section~\ref{sec:non-finish}) $r_{a}$ 
    \item One-time goal-reaching reward $r_g$
    \item Constant collision penalty for each frame when an agent collides with another agent or an obstacle $r_c$  
\end{enumerate}

A complete reward function is a weighted sum of a subset of these terms. For terms (1)-(4) and (8)-(9), their coefficients are equal to $1$ due to their physics-based formulation. Terms (1)-(3) and (5)-(7) are also multiplied by the duration of the timestep in the simulation.

We primarily focus on evaluating the following reward functions. Note that all of these variants include components (10) and (11) (goal-reaching and collision penalty, respectively)

\begin{enumerate}[label=(\alph*)]
    \item \textbf{Base curriculum} -- a curriculum which initially has components (4), (6) (with $c_e = 2$), and after 200 training steps, switches to (1), (3), (4), (9)
    \item \textbf{Base curriculum (no acceleration)} -- like (a), but using component (2) instead of (3)
    \item \textbf{Base curriculum (no heuristic)} -- like (a), but without component (9)
    \item \textbf{Base curriculum (optimal heuristic)} -- like (a), but using component (8) instead of (9)
    \item \textbf{Energy (acceleration)} -- components (1), (3), (4)
    \item \textbf{Energy (no acceleration)} -- components (1), (2), (4)
    \item \textbf{Energy (no potential)} -- components (1), (2)
    \item \textbf{Speed matching} -- components (4), (5), based on \citet{kwiatkowski_understanding_2023}
    \item \textbf{Speeding penalty} -- components (4), (6), based on \citet{lee_crowd_2018}
    \item \textbf{Exponential velocity matching} -- component (7), based on \citet{xu_human-inspired_2021}
\end{enumerate}

We trained agents using each of these reward functions, and summarize the results in Section~\ref{sec:results}.

Furthermore, to investigate the importance of the potential term, we also evaluated the following reward functions:
\begin{enumerate}[label=(\Alph*)]
    \item \textbf{Base curriculum} -- same as reward (a), serving as a baseline
    \item \textbf{No potential} -- same as (A), but without component (4) (potential)
    \item \textbf{No potential and final penalty} -- same as (B), but also without component (9) (non-finishing penalty)
    \item \textbf{No potential and goal} -- same as (B), but also without component (10)
    \item \textbf{No potential and goal, optimal heuristic} -- same as (D), but with component (10) instead of (9)
    \item \textbf{Pure energy} -- same as (C), but also without component (10) (goal). The second phase of the curriculum only uses components (1) and (3).
    \item \textbf{Pure energy, no discounting} -- same as (F), but the discount factor is set to $\gamma = 1$ throughout the training
\end{enumerate}
We describe the results of these experiments in Section~\ref{sec:ispotential}.

\section{Results}\label{sec:results}

\begin{table*}[]
\caption{Mean value of the Energy+ metric after training in a given scenario, using a given reward function. Each value is based on 8 independent training runs. Lower is better}
\label{tab:base-results}
\begin{tabular}{|c|c|c|c|c|c|c}
\cline{1-6}
\textbf{} &
  \textbf{Circle} &
  \textbf{Crossing} &
  \textbf{Corridor} &
  \textbf{Car} &
  \textbf{Choke} &
   \\ \cline{1-6}
\textbf{Base curriculum} &
  \textbf{58.2 ± 0.54} &
  66.38 ± 1.39 &
  77.56 ± 6.19 &
  110.95 ± 3.99 &
  94.97 ± 4.03 &
  \begin{tabular}[c]{@{}c@{}}\phantom{l}\\ \phantom{l}\end{tabular} \\ \cline{1-6}
\textbf{\begin{tabular}[c]{@{}c@{}}Base curriculum\\ (no acceleration)\end{tabular}} &
  61.62 ± 0.82 &
  72.56 ± 1.14 &
  85.26 ± 6.07 &
  112.81 ± 2.51 &
  112.18 ± 5.49 &
  \begin{tabular}[c]{@{}c@{}}\phantom{l}\\ \phantom{l}\end{tabular} \\ \cline{1-6}
\textbf{\begin{tabular}[c]{@{}c@{}}Base curriculum\\ (no heuristic)\end{tabular}} &
  59.18 ± 0.51 &
  \textbf{65.81 ± 1.03} &
  \textbf{63.29 ± 0.32} &
  95.63 ± 8.31 &
  114.78 ± 12.55 &
  \begin{tabular}[c]{@{}c@{}}\phantom{l}\\ \phantom{l}\end{tabular} \\ \cline{1-6}
\textbf{\begin{tabular}[c]{@{}c@{}}Base curriculum\\ (optimal heuristic)\end{tabular}} &
  59.17 ± 1.01 &
  67.56 ± 2.12 &
  69.34 ± 2.1 &
  103.76 ± 6.57 &
  \textbf{94.53 ± 7.99} &
  \begin{tabular}[c]{@{}c@{}}\phantom{l}\\ \phantom{l}\end{tabular} \\ \cline{1-6}
\textbf{\begin{tabular}[c]{@{}c@{}}Energy\\ (acceleration)\end{tabular}} &
  74.59 ± 2.48 &
  73.55 ± 3.36 &
  96.19 ± 9.02 &
  \textbf{85.05 ± 9.36} &
  105.58 ± 9.09 &
  \begin{tabular}[c]{@{}c@{}}\phantom{l}\\ \phantom{l}\end{tabular} \\ \cline{1-6}
\textbf{\begin{tabular}[c]{@{}c@{}}Energy\\ (no acceleration)\end{tabular}} &
  67.1 ± 1.97 &
  81.32 ± 3.26 &
  102.75 ± 5.08 &
  108.32 ± 1.26 &
  106.39 ± 5.43 &
  \begin{tabular}[c]{@{}c@{}}\phantom{l}\\ \phantom{l}\end{tabular} \\ \cline{1-6}
\textbf{\begin{tabular}[c]{@{}c@{}}Energy\\ (no potential)\end{tabular}} &
  459.53 ± 53.16 &
  454.01 ± 5.29 &
  450.65 ± 5.06 &
  463.26 ± 1.44 &
  460.28 ± 1.0 &
  \begin{tabular}[c]{@{}c@{}}\phantom{l}\\ \phantom{l}\end{tabular} \\ \cline{1-6}
\textbf{Speed matching} &
  60.13 ± 0.71 &
  81.04 ± 5.66 &
  68.35 ± 1.7 &
  126.28 ± 2.03 &
  276.55 ± 14.99 &
  \begin{tabular}[c]{@{}c@{}}\phantom{l}\\ \phantom{l}\end{tabular} \\ \cline{1-6}
\textbf{Speeding penalty} &
  58.55 ± 0.87 &
  88.47 ± 4.42 &
  98.71 ± 6.06 &
  119.72 ± 1.27 &
  130.03 ± 5.66 &
  \begin{tabular}[c]{@{}c@{}}\phantom{l}\\ \phantom{l}\end{tabular} \\ \cline{1-6}
\textbf{\begin{tabular}[c]{@{}c@{}}Exponential\\ velocity matching\end{tabular}} &
  63.5 ± 1.73 &
  77.18 ± 4.33 &
  85.33 ± 4.91 &
  107.66 ± 0.7 &
  129.9 ± 9.4 &
  \begin{tabular}[c]{@{}c@{}}\phantom{l}\\ \phantom{l}\end{tabular} \\ \cline{1-6}
\end{tabular}
\end{table*}

\begin{figure}
    \centering
    \includegraphics[width=\linewidth]{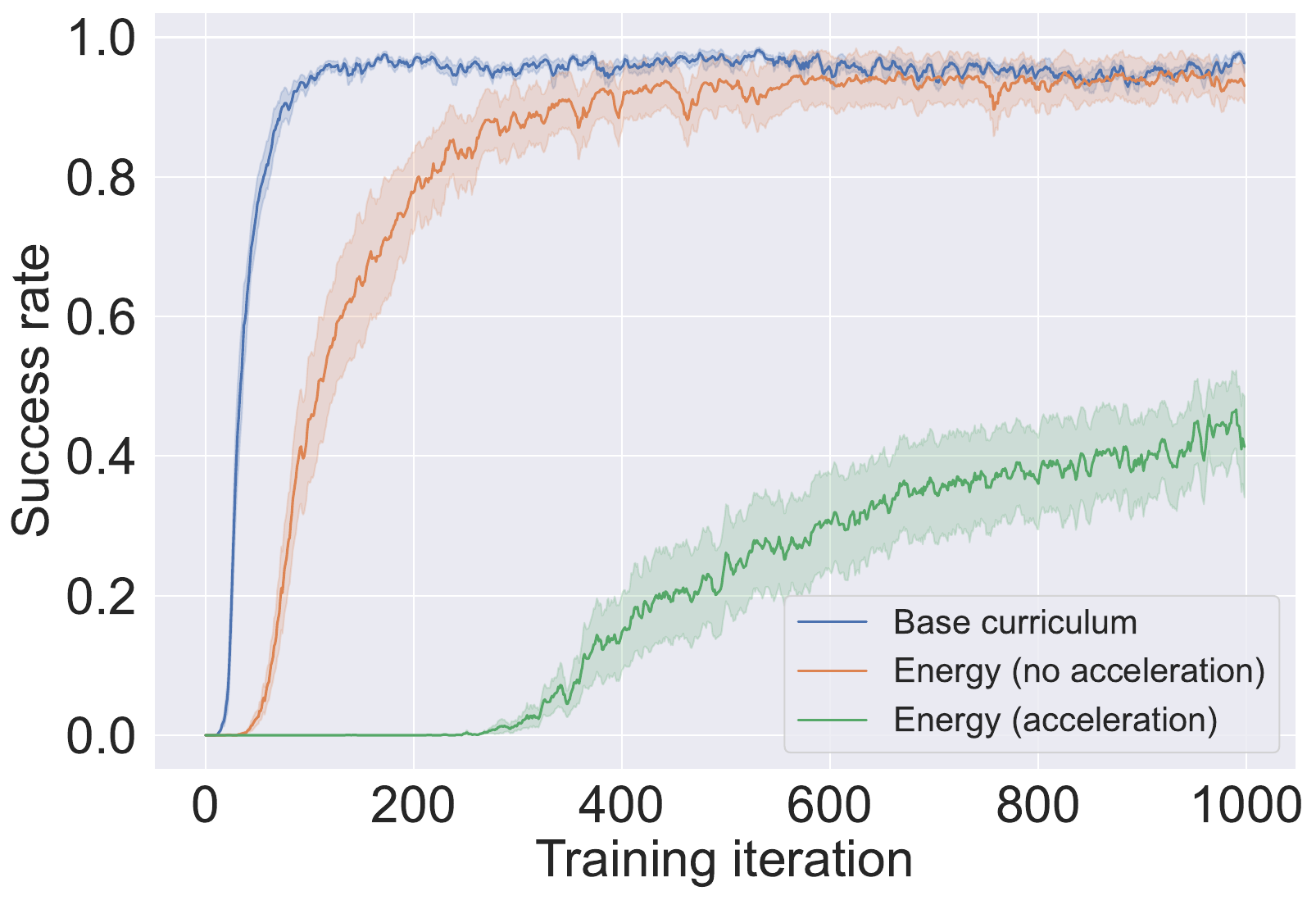}
    \caption{Success rates of agents trained with certain reward functions in the Circle scenario.}
    \label{fig:success-rate}
\end{figure}

While the details differ based on the scenario, in all of them except for the Car scenario, the best-performing reward is a curriculum leading to energy optimization. In the Car scenario, the best-performing reward in terms of the Energy+ metric is directly optimizing energy from the beginning.

The benefit of the curriculum becomes apparent when we consider the progression of the training. We show the success rates in the Circle scenario as a function of the training steps in Figure~\ref{fig:success-rate}. This scenario has a difficult coordination task embedded in it -- when agents travel through the central part of the scene, they must avoid many other agents moving in all directions to prevent collisions. Each collision may lead to additional energy usage in order to resume movement, which effectively increases the collision penalty. Because of this, agents learn the navigation task much more slowly. Conversely, using a simple speeding penalty for the initial part of the training allows the agents to quickly reach a high success rate, which is then maintained after the reward is switched to energy optimization.

On the other hand, in the Car scenario, the best-performing variant is direct energy optimization. This is because agents trained with speeding penalty (as opposed to energy minimization) initially converge to attempting to quickly go in front of the car, passing before it hits them. In contrast, agents trained to minimize energy usage simply wait for the obstacle to pass, or start moving behind it. It is difficult to progressively switch from the former to the latter behavior, so the curriculum fails to produce efficient behavior.

\subsection{Is potential necessary?}\label{sec:ispotential}

\begin{figure}
    \centering
    \includegraphics[width=\linewidth]{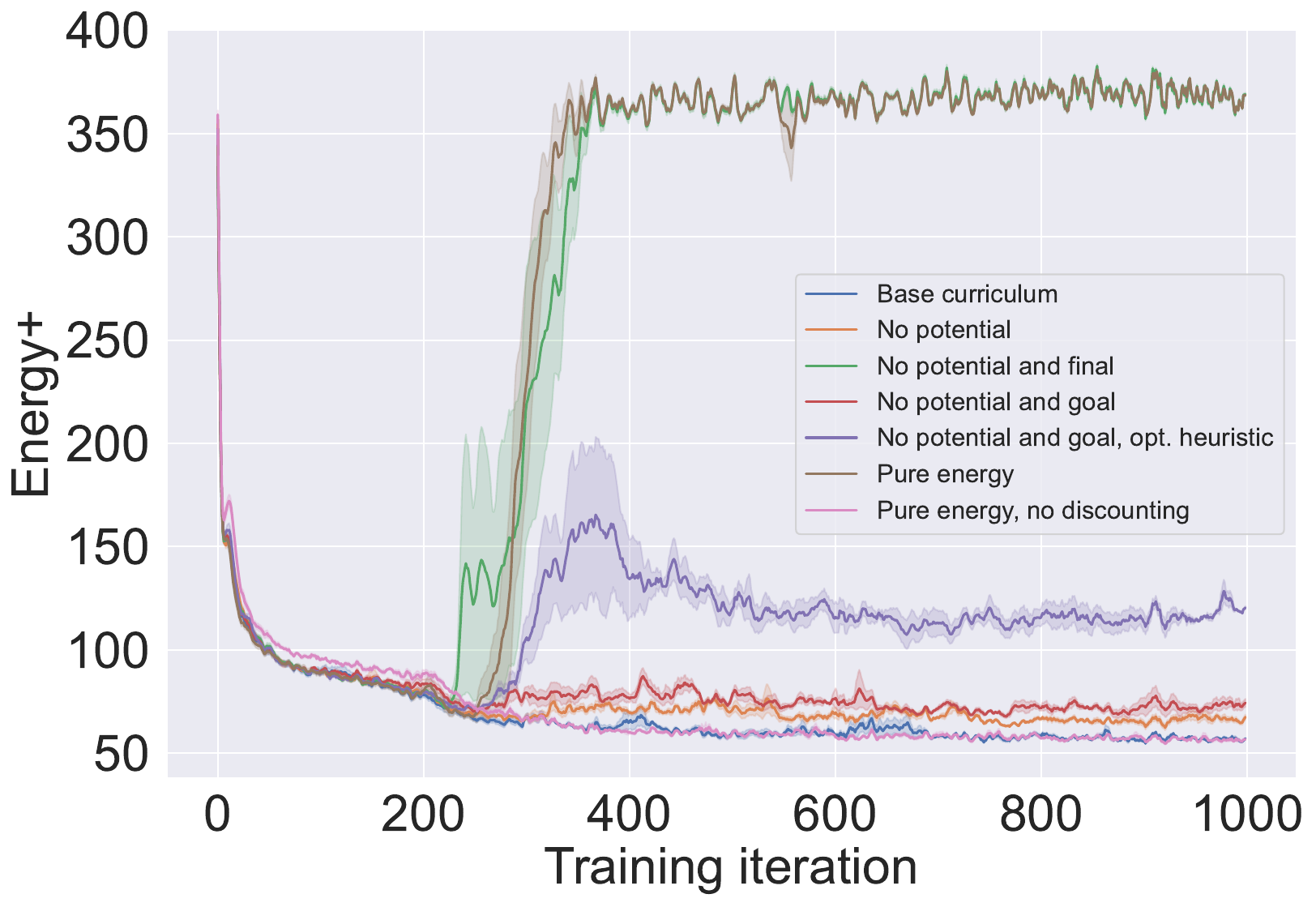}
    \caption{Energy+ metric as a function of training progress with various reward functions. To maintain the performance from the first stage of the training, it is necessary to either use a potential term, or set the discount factor to $\gamma=1$. Agents without a potential or a final heuristic converge to standing still, while other variants' performance significantly degrades.}
    \label{fig:pureenergy}
\end{figure}

In Section~\ref{sec:energy-reward}, we provide theoretical justification for why simply optimizing energy is likely to fail. The data in Table~\ref{tab:base-results} confirms at least the local optimum argument -- directly optimizing the energy usage consistently leads to the worst performance, corresponding to standing still.
To empirically validate our global optimum argument, we conducted additional experiments on the Circle scenario, using reward functions (A)-(G).

We show the results in terms of the Energy+ values in Figure~\ref{fig:pureenergy}. The \textbf{no potential} variant maintains a reasonable performance, but its energy efficiency drops compared to the baseline. Both variants without the final non-finishing penalty (with or without the goal reward -- (C) and (F) respectively) rapidly deteriorate to a policy which stays still for the entire duration of the episode. The variants that retain some of their performance are (B) and (D), \ie ones which still use the average heuristic penalty for not reaching the goal, however their success rate is significantly lower than the baseline. Using the optimal heuristic (E) instead of the average heuristic degrades performance significantly, leading agents to slowly approach the goal, abusing the generous reward they receive at the end of the episode. Finally, using pure energy optimization in a curriculum without a discount factor retains the same performance as the base curriculum.

This confirms that absent of additional goals, with a discount factor of $\gamma=0.99$, using energy as a reward without a guiding potential fails to converge to a valid policy, even when initialized with a goal-seeking policy trained with a different reward function. This may be mitigated by including a guiding potential, which in some cases enables effective end-to-end training using that reward function. Alternatively, if the training converges without discounting, \ie with $\gamma = 1$, pure energy may also be a valid approach as a second (or later) stage of a curriculum. This is consistent with the analysis by \citet{naik_discounted_2019}, who describe theoretical problems with the discounted utility paradigm.

\subsection{Impact of acceleration}

\begin{figure}
    \centering
    \includegraphics[width=\linewidth]{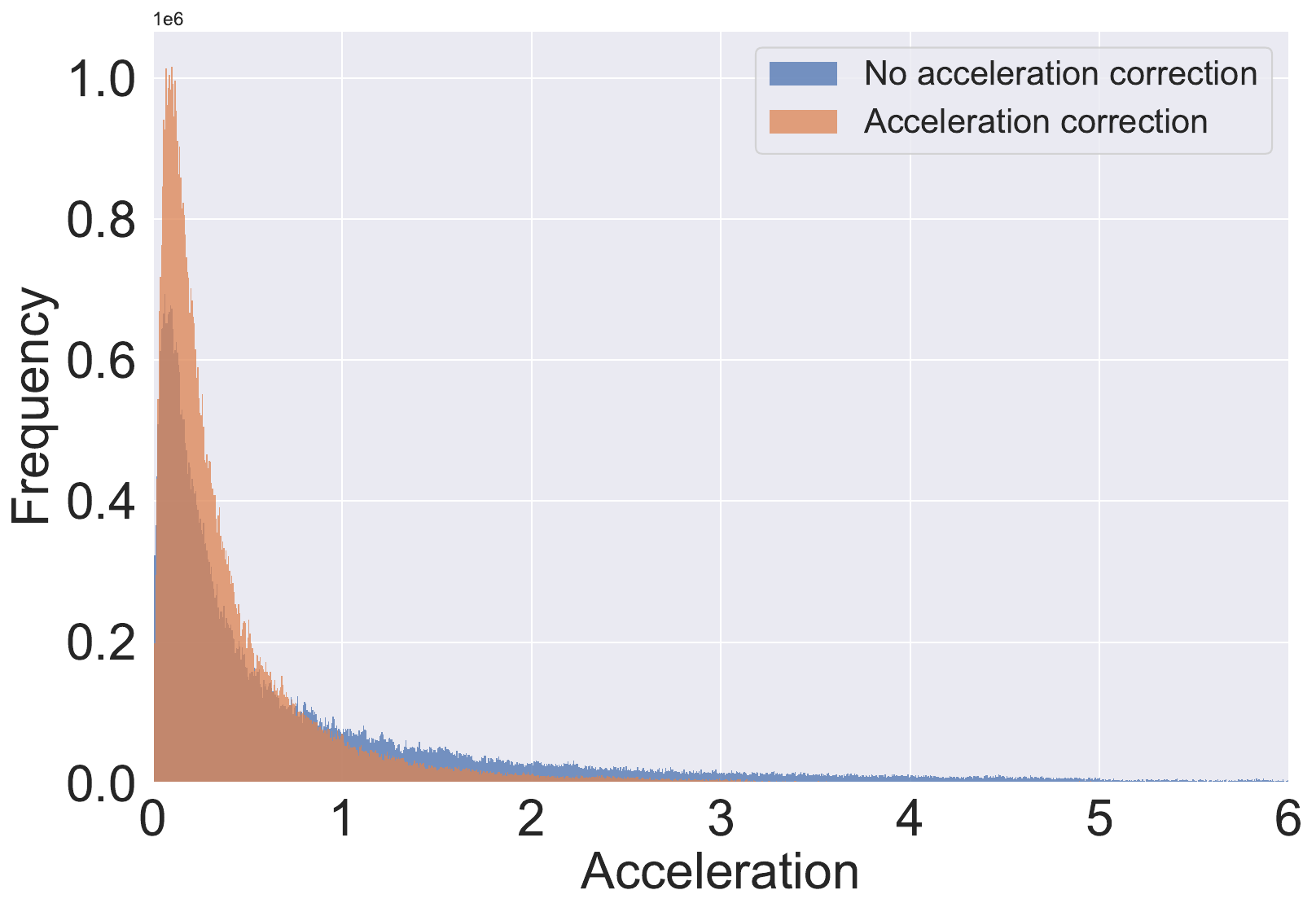}
    \caption{Histogram of accelerations in the Circle scenario, trained with and without the acceleration-based term in the reward function.}
    \label{fig:acceleration-histogram}
\end{figure}

In order to evaluate the impact of the acceleration correction to the energy estimation introduced in Section~\ref{sec:energy-acceleration}, we compare agents trained with the base curriculum, with and without the acceleration correction. We show the histogram of accelerations, collected across 8 independent training runs in the Circle scenario, in Figure~\ref{fig:acceleration-histogram}.

The average magnitude of the acceleration across the trajectories is \SI{0.339}{\meter/\second^2} with the acceleration correction in the energy estimation, and \SI{0.679}{\meter/\second^2} without it. This result is statistically significant with $p < 0.01$ using the two-sample Kolmogorov-Smirnov test. This shows that including the acceleration in energy estimation successfully leads to smoother behavior. At the same time, the energy usage without the acceleration correction remains similar for both variants -- $49.77 \pm 1.077$ and $50.47 \pm 1.284$ respectively, indicating that the reduced acceleration does not come at the cost of otherwise less efficient movement. 


\section{Conclusions}

In this work, we introduce two contributions: a new, more accurate way to estimate energy usage in the context of crowd simulation, and a novel reward function formulation for training agents navigating in an energy-efficient manner. We demonstrate a successful curriculum learning approach, where an initial speeding penalty is replaced by a simpler energy optimization formulation in later training stages. This method allows the agents to learn basic navigation first, and then focus on efficiency. 

Our experiments on several crowd navigation scenarios show that training using an energy-based reward consistently outperform other reward functions used in prior work. A critical component of this reward structure is the guiding potential, which ensures that agents navigate towards the goal, and do not simply stay still to minimize the energy usage. We empirically verify this conclusion through additional experiments that exclude this term from the reward function.

Interestingly, in some scenarios, such as the Car scenario, a curriculum approach does not provide any benefits, and agents perform optimally when trained directly with the energy optimization reward. This can be attributed to the specific nature of this scenario, where the initial policy learned by the agents with a speeding penalty makes them rush in front of the moving obstacle, a strategy that contrasts with the more efficient wait-and-follow behavior learned through energy optimization. This highlights the potential need for a more scenario-specific reward formulation or a flexible curriculum training approach that can adjust itself based on the scenario complexity and nature.

Furthermore, our analysis of discount factor effects on training outcomes with a pure energy reward function aligns with the theoretical discussions raised by \citet{naik_discounted_2019}. It shows that if the training is conducted without discounting, using energy as a reward without a guiding potential can converge to a valid policy when initialized with a goal-seeking policy trained with a different reward function. This discovery invites future research to explore the utility of different discounting paradigms in such energy optimization tasks and potentially other reinforcement learning applications. 

While the current results are promising, several directions remain for future work. The energy estimation for motion with acceleration could be made more accurate by considering the agent's physical model more closely. Additionally, the potential function could be replaced with a more sophisticated heuristic that considers the actual shortest path to the goal, taking into account other agents and obstacles. Another possible direction could be developing an adaptive curriculum that consider the nature of the scenario or the learning progress of the agent. Finally, integrating this energy-efficient approach with social norms and considering more realistic crowd behaviors could lead to generating more realistic behaviors with RL.

\bibliographystyle{ACM-Reference-Format}
\bibliography{references}


\begin{thebibliography}{29}


\ifx \showCODEN    \undefined \def \showCODEN     #1{\unskip}     \fi
\ifx \showDOI      \undefined \def \showDOI       #1{#1}\fi
\ifx \showISBNx    \undefined \def \showISBNx     #1{\unskip}     \fi
\ifx \showISBNxiii \undefined \def \showISBNxiii  #1{\unskip}     \fi
\ifx \showISSN     \undefined \def \showISSN      #1{\unskip}     \fi
\ifx \showLCCN     \undefined \def \showLCCN      #1{\unskip}     \fi
\ifx \shownote     \undefined \def \shownote      #1{#1}          \fi
\ifx \showarticletitle \undefined \def \showarticletitle #1{#1}   \fi
\ifx \showURL      \undefined \def \showURL       {\relax}        \fi
\providecommand\bibfield[2]{#2}
\providecommand\bibinfo[2]{#2}
\providecommand\natexlab[1]{#1}
\providecommand\showeprint[2][]{arXiv:#2}

\bibitem[Abbeel and Ng(2004)]%
        {abbeel_apprenticeship_2004}
\bibfield{author}{\bibinfo{person}{Pieter Abbeel} {and} \bibinfo{person}{Andrew~Y. Ng}.} \bibinfo{year}{2004}\natexlab{}.
\newblock \showarticletitle{Apprenticeship learning via inverse reinforcement learning}. In \bibinfo{booktitle}{\emph{Proceedings of the twenty-first international conference on {Machine} learning}} \emph{(\bibinfo{series}{{ICML} '04})}. \bibinfo{publisher}{Association for Computing Machinery}, \bibinfo{address}{New York, NY, USA}, \bibinfo{pages}{1}.
\newblock
\showISBNx{978-1-58113-838-2}
\urldef\tempurl%
\url{https://doi.org/10.1145/1015330.1015430}
\showDOI{\tempurl}


\bibitem[Andrychowicz et~al\mbox{.}(2020)]%
        {andrychowicz_what_2020}
\bibfield{author}{\bibinfo{person}{Marcin Andrychowicz}, \bibinfo{person}{Anton Raichuk}, \bibinfo{person}{Piotr Stańczyk}, \bibinfo{person}{Manu Orsini}, \bibinfo{person}{Sertan Girgin}, \bibinfo{person}{Raphael Marinier}, \bibinfo{person}{Léonard Hussenot}, \bibinfo{person}{Matthieu Geist}, \bibinfo{person}{Olivier Pietquin}, \bibinfo{person}{Marcin Michalski}, \bibinfo{person}{Sylvain Gelly}, {and} \bibinfo{person}{Olivier Bachem}.} \bibinfo{year}{2020}\natexlab{}.
\newblock \showarticletitle{What {Matters} {In} {On}-{Policy} {Reinforcement} {Learning}? {A} {Large}-{Scale} {Empirical} {Study}}.
\newblock \bibinfo{journal}{\emph{arXiv:2006.05990 [cs, stat]}} (\bibinfo{date}{June} \bibinfo{year}{2020}).
\newblock
\urldef\tempurl%
\url{http://arxiv.org/abs/2006.05990}
\showURL{%
\tempurl}
\newblock
\shownote{arXiv: 2006.05990}.


\bibitem[Bruneau et~al\mbox{.}(2015)]%
        {bruneau_going_2015}
\bibfield{author}{\bibinfo{person}{Julien Bruneau}, \bibinfo{person}{Anne-Hélène Olivier}, {and} \bibinfo{person}{Julien Pettré}.} \bibinfo{year}{2015}\natexlab{}.
\newblock \showarticletitle{Going {Through}, {Going} {Around}: {A} {Study} on {Individual} {Avoidance} of {Groups}}.
\newblock \bibinfo{journal}{\emph{IEEE Transactions on Visualization and Computer Graphics}} \bibinfo{volume}{21}, \bibinfo{number}{4} (\bibinfo{date}{April} \bibinfo{year}{2015}), \bibinfo{pages}{9}.
\newblock
\urldef\tempurl%
\url{https://doi.org/10.1109/TVCG.2015.2391862}
\showDOI{\tempurl}


\bibitem[Engstrom et~al\mbox{.}(2020)]%
        {engstrom_implementation_2020}
\bibfield{author}{\bibinfo{person}{Logan Engstrom}, \bibinfo{person}{Andrew Ilyas}, \bibinfo{person}{Shibani Santurkar}, \bibinfo{person}{Dimitris Tsipras}, \bibinfo{person}{Firdaus Janoos}, \bibinfo{person}{Larry Rudolph}, {and} \bibinfo{person}{Aleksander Madry}.} \bibinfo{year}{2020}\natexlab{}.
\newblock \showarticletitle{Implementation {Matters} in {Deep} {Policy} {Gradients}: {A} {Case} {Study} on {PPO} and {TRPO}}.
\newblock \bibinfo{journal}{\emph{arXiv:2005.12729 [cs, stat]}} (\bibinfo{date}{May} \bibinfo{year}{2020}).
\newblock
\urldef\tempurl%
\url{http://arxiv.org/abs/2005.12729}
\showURL{%
\tempurl}
\newblock
\shownote{arXiv: 2005.12729}.


\bibitem[Fedus et~al\mbox{.}(2019)]%
        {fedus_hyperbolic_2019}
\bibfield{author}{\bibinfo{person}{William Fedus}, \bibinfo{person}{Carles Gelada}, \bibinfo{person}{Yoshua Bengio}, \bibinfo{person}{Marc~G. Bellemare}, {and} \bibinfo{person}{Hugo Larochelle}.} \bibinfo{year}{2019}\natexlab{}.
\newblock \showarticletitle{Hyperbolic {Discounting} and {Learning} over {Multiple} {Horizons}}.
\newblock \bibinfo{journal}{\emph{arXiv:1902.06865 [cs, stat]}} (\bibinfo{date}{Feb.} \bibinfo{year}{2019}).
\newblock
\newblock
\shownote{arXiv: 1902.06865}.


\bibitem[Guy et~al\mbox{.}(2010)]%
        {guy_pledestrians_2010}
\bibfield{author}{\bibinfo{person}{Stephen~J. Guy}, \bibinfo{person}{Jatin Chhugani}, \bibinfo{person}{Sean Curtis}, \bibinfo{person}{Pradeep Dubey}, \bibinfo{person}{Ming Lin}, {and} \bibinfo{person}{Dinesh Manocha}.} \bibinfo{year}{2010}\natexlab{}.
\newblock \showarticletitle{{PLEdestrians}: {A} {Least}-{Effort} {Approach} to {Crowd} {Simulation}}.
\newblock \bibinfo{journal}{\emph{Eurographics/ ACM SIGGRAPH Symposium on Computer Animation}} (\bibinfo{year}{2010}), \bibinfo{pages}{10 pages}.
\newblock
\showISSN{1727-5288}
\urldef\tempurl%
\url{https://doi.org/10.2312/SCA/SCA10/119-128}
\showDOI{\tempurl}
\newblock
\shownote{Artwork Size: 10 pages ISBN: 9783905674279 Publisher: The Eurographics Association}.


\bibitem[Hu et~al\mbox{.}(2022)]%
        {hu_heterogeneous_2022}
\bibfield{author}{\bibinfo{person}{Kaidong Hu}, \bibinfo{person}{Michael~Brandon Haworth}, \bibinfo{person}{Glen Berseth}, \bibinfo{person}{Vladimir Pavlovic}, \bibinfo{person}{Petros Faloutsos}, {and} \bibinfo{person}{Mubbasir Kapadia}.} \bibinfo{year}{2022}\natexlab{}.
\newblock \showarticletitle{Heterogeneous {Crowd} {Simulation} using {Parametric} {Reinforcement} {Learning}}.
\newblock \bibinfo{journal}{\emph{IEEE Transactions on Visualization and Computer Graphics}} (\bibinfo{year}{2022}), \bibinfo{pages}{1--1}.
\newblock
\showISSN{1077-2626, 1941-0506, 2160-9306}
\urldef\tempurl%
\url{https://doi.org/10.1109/TVCG.2021.3139031}
\showDOI{\tempurl}


\bibitem[Huang et~al\mbox{.}(2022)]%
        {huang_37_2022}
\bibfield{author}{\bibinfo{person}{Shengyi Huang}, \bibinfo{person}{Rousslan Fernand~Julien Dossa}, \bibinfo{person}{Antonin Raffin}, \bibinfo{person}{Anssi Kanervisto}, {and} \bibinfo{person}{Weixun Wang}.} \bibinfo{year}{2022}\natexlab{}.
\newblock \showarticletitle{The 37 {Implementation} {Details} of {Proximal} {Policy} {Optimization}}. In \bibinfo{booktitle}{\emph{{ICLR} {Blog} {Track}}}.
\newblock
\urldef\tempurl%
\url{https://iclr-blog-track.github.io/2022/03/25/ppo-implementation-details/}
\showURL{%
\tempurl}


\bibitem[Kwiatkowski et~al\mbox{.}(2022)]%
        {kwiatkowski_survey_2022}
\bibfield{author}{\bibinfo{person}{Ariel Kwiatkowski}, \bibinfo{person}{Eduardo Alvarado}, \bibinfo{person}{Vicky Kalogeiton}, \bibinfo{person}{C.~Karen Liu}, \bibinfo{person}{Julien Pettré}, \bibinfo{person}{Michiel van~de Panne}, {and} \bibinfo{person}{Marie‐Paule Cani}.} \bibinfo{year}{2022}\natexlab{}.
\newblock \showarticletitle{A {Survey} on {Reinforcement} {Learning} {Methods} in {Character} {Animation}}.
\newblock \bibinfo{journal}{\emph{Computer Graphics Forum}} \bibinfo{volume}{41}, \bibinfo{number}{2} (\bibinfo{date}{May} \bibinfo{year}{2022}), \bibinfo{pages}{613--639}.
\newblock
\showISSN{0167-7055, 1467-8659}
\urldef\tempurl%
\url{https://doi.org/10.1111/cgf.14504}
\showDOI{\tempurl}


\bibitem[Kwiatkowski et~al\mbox{.}(2023a)]%
        {kwiatkowski_ugae_2023}
\bibfield{author}{\bibinfo{person}{Ariel Kwiatkowski}, \bibinfo{person}{Vicky Kalogeiton}, \bibinfo{person}{Julien Pettré}, {and} \bibinfo{person}{Marie-Paule Cani}.} \bibinfo{year}{2023}\natexlab{a}.
\newblock \bibinfo{title}{{UGAE}: {A} {Novel} {Approach} to {Non}-exponential {Discounting}}.
\newblock
\newblock
\urldef\tempurl%
\url{https://doi.org/10.48550/arXiv.2302.05740}
\showDOI{\tempurl}
\newblock
\shownote{arXiv:2302.05740 [cs]}.


\bibitem[Kwiatkowski et~al\mbox{.}(2023b)]%
        {kwiatkowski_understanding_2023}
\bibfield{author}{\bibinfo{person}{Ariel Kwiatkowski}, \bibinfo{person}{Vicky Kalogeiton}, \bibinfo{person}{Julien Pettré}, {and} \bibinfo{person}{Marie-Paule Cani}.} \bibinfo{year}{2023}\natexlab{b}.
\newblock \showarticletitle{Understanding reinforcement learned crowds}.
\newblock \bibinfo{journal}{\emph{Computers \& Graphics}}  \bibinfo{volume}{110} (\bibinfo{date}{Feb.} \bibinfo{year}{2023}), \bibinfo{pages}{28--37}.
\newblock
\showISSN{00978493}
\urldef\tempurl%
\url{https://doi.org/10.1016/j.cag.2022.11.007}
\showDOI{\tempurl}


\bibitem[Lee et~al\mbox{.}(2018)]%
        {lee_crowd_2018}
\bibfield{author}{\bibinfo{person}{Jaedong Lee}, \bibinfo{person}{Jungdam Won}, {and} \bibinfo{person}{Jehee Lee}.} \bibinfo{year}{2018}\natexlab{}.
\newblock \showarticletitle{Crowd simulation by deep reinforcement learning}. In \bibinfo{booktitle}{\emph{Proceedings of the 11th {Annual} {International} {Conference} on {Motion}, {Interaction}, and {Games}}}. \bibinfo{publisher}{ACM}, \bibinfo{address}{Limassol Cyprus}, \bibinfo{pages}{1--7}.
\newblock
\showISBNx{978-1-4503-6015-9}
\urldef\tempurl%
\url{https://doi.org/10.1145/3274247.3274510}
\showDOI{\tempurl}


\bibitem[Leibo et~al\mbox{.}(2017)]%
        {leibo_multi-agent_2017}
\bibfield{author}{\bibinfo{person}{Joel~Z. Leibo}, \bibinfo{person}{Vinicius Zambaldi}, \bibinfo{person}{Marc Lanctot}, \bibinfo{person}{Janusz Marecki}, {and} \bibinfo{person}{Thore Graepel}.} \bibinfo{year}{2017}\natexlab{}.
\newblock \showarticletitle{Multi-{Agent} {Reinforcement} {Learning} in {Sequential} {Social} {Dilemmas}}. In \bibinfo{booktitle}{\emph{Proceedings of the 16th {Conference} on {Autonomous} {Agents} and {MultiAgent} {Systems}}} \emph{(\bibinfo{series}{{AAMAS} '17})}. \bibinfo{publisher}{International Foundation for Autonomous Agents and Multiagent Systems}, \bibinfo{address}{Richland, SC}, \bibinfo{pages}{464--473}.
\newblock
\newblock
\shownote{event-place: São Paulo, Brazil}.


\bibitem[Long et~al\mbox{.}(2018)]%
        {long_towards_2018}
\bibfield{author}{\bibinfo{person}{Pinxin Long}, \bibinfo{person}{Tingxiang Fan}, \bibinfo{person}{Xinyi Liao}, \bibinfo{person}{Wenxi Liu}, \bibinfo{person}{Hao Zhang}, {and} \bibinfo{person}{Jia Pan}.} \bibinfo{year}{2018}\natexlab{}.
\newblock \showarticletitle{Towards {Optimally} {Decentralized} {Multi}-{Robot} {Collision} {Avoidance} via {Deep} {Reinforcement} {Learning}}.
\newblock \bibinfo{journal}{\emph{arXiv:1709.10082 [cs]}} (\bibinfo{date}{May} \bibinfo{year}{2018}).
\newblock
\newblock
\shownote{arXiv: 1709.10082}.


\bibitem[Lv et~al\mbox{.}(2022)]%
        {lv_emotional_2022}
\bibfield{author}{\bibinfo{person}{Pei Lv}, \bibinfo{person}{Qingqing Yu}, \bibinfo{person}{Boya Xu}, \bibinfo{person}{Chaochao Li}, \bibinfo{person}{Bing Zhou}, {and} \bibinfo{person}{Mingliang Xu}.} \bibinfo{year}{2022}\natexlab{}.
\newblock \bibinfo{title}{Emotional {Contagion}-{Aware} {Deep} {Reinforcement} {Learning} for {Antagonistic} {Crowd} {Simulation}}.
\newblock
\newblock
\urldef\tempurl%
\url{https://doi.org/10.48550/arXiv.2105.00854}
\showDOI{\tempurl}
\newblock
\shownote{arXiv:2105.00854 [physics]}.


\bibitem[Naik et~al\mbox{.}(2019)]%
        {naik_discounted_2019}
\bibfield{author}{\bibinfo{person}{Abhishek Naik}, \bibinfo{person}{Roshan Shariff}, \bibinfo{person}{Niko Yasui}, \bibinfo{person}{Hengshuai Yao}, {and} \bibinfo{person}{Richard~S. Sutton}.} \bibinfo{year}{2019}\natexlab{}.
\newblock \showarticletitle{Discounted {Reinforcement} {Learning} {Is} {Not} an {Optimization} {Problem}}.
\newblock \bibinfo{journal}{\emph{arXiv:1910.02140 [cs]}} (\bibinfo{date}{Nov.} \bibinfo{year}{2019}).
\newblock
\newblock
\shownote{arXiv: 1910.02140}.


\bibitem[Ng et~al\mbox{.}(1999)]%
        {ng_policy_1999}
\bibfield{author}{\bibinfo{person}{Andrew~Y. Ng}, \bibinfo{person}{Daishi Harada}, {and} \bibinfo{person}{Stuart~J. Russell}.} \bibinfo{year}{1999}\natexlab{}.
\newblock \showarticletitle{Policy {Invariance} {Under} {Reward} {Transformations}: {Theory} and {Application} to {Reward} {Shaping}}. In \bibinfo{booktitle}{\emph{Proceedings of the {Sixteenth} {International} {Conference} on {Machine} {Learning}}} \emph{(\bibinfo{series}{{ICML} '99})}. \bibinfo{publisher}{Morgan Kaufmann Publishers Inc.}, \bibinfo{address}{San Francisco, CA, USA}, \bibinfo{pages}{278--287}.
\newblock
\showISBNx{1-55860-612-2}


\bibitem[Ng and Russell(2000)]%
        {ng_algorithms_2000}
\bibfield{author}{\bibinfo{person}{Andrew~Y. Ng} {and} \bibinfo{person}{Stuart~J. Russell}.} \bibinfo{year}{2000}\natexlab{}.
\newblock \showarticletitle{Algorithms for {Inverse} {Reinforcement} {Learning}}. In \bibinfo{booktitle}{\emph{Proceedings of the {Seventeenth} {International} {Conference} on {Machine} {Learning}}} \emph{(\bibinfo{series}{{ICML} '00})}. \bibinfo{publisher}{Morgan Kaufmann Publishers Inc.}, \bibinfo{address}{San Francisco, CA, USA}, \bibinfo{pages}{663--670}.
\newblock
\showISBNx{978-1-55860-707-1}


\bibitem[Panayiotou et~al\mbox{.}(2022)]%
        {panayiotou_ccp_2022}
\bibfield{author}{\bibinfo{person}{Andreas Panayiotou}, \bibinfo{person}{Theodoros Kyriakou}, \bibinfo{person}{Marilena Lemonari}, \bibinfo{person}{Yiorgos Chrysanthou}, {and} \bibinfo{person}{Panayiotis Charalambous}.} \bibinfo{year}{2022}\natexlab{}.
\newblock \showarticletitle{{CCP}: {Configurable} {Crowd} {Profiles}}. In \bibinfo{booktitle}{\emph{Special {Interest} {Group} on {Computer} {Graphics} and {Interactive} {Techniques} {Conference} {Proceedings}}}. \bibinfo{publisher}{ACM}, \bibinfo{address}{Vancouver BC Canada}, \bibinfo{pages}{1--10}.
\newblock
\showISBNx{978-1-4503-9337-9}
\urldef\tempurl%
\url{https://doi.org/10.1145/3528233.3530712}
\showDOI{\tempurl}


\bibitem[Schulman et~al\mbox{.}(2017)]%
        {schulman_proximal_2017}
\bibfield{author}{\bibinfo{person}{John Schulman}, \bibinfo{person}{Filip Wolski}, \bibinfo{person}{Prafulla Dhariwal}, \bibinfo{person}{Alec Radford}, {and} \bibinfo{person}{Oleg Klimov}.} \bibinfo{year}{2017}\natexlab{}.
\newblock \showarticletitle{Proximal {Policy} {Optimization} {Algorithms}}.
\newblock \bibinfo{journal}{\emph{arXiv:1707.06347 [cs]}} (\bibinfo{date}{Aug.} \bibinfo{year}{2017}).
\newblock


\bibitem[Sun et~al\mbox{.}(2019)]%
        {sun_crowd_2019}
\bibfield{author}{\bibinfo{person}{L. Sun}, \bibinfo{person}{J. Zhai}, {and} \bibinfo{person}{W. Qin}.} \bibinfo{year}{2019}\natexlab{}.
\newblock \showarticletitle{Crowd {Navigation} in an {Unknown} and {Dynamic} {Environment} {Based} on {Deep} {Reinforcement} {Learning}}.
\newblock \bibinfo{journal}{\emph{IEEE Access}}  \bibinfo{volume}{7} (\bibinfo{year}{2019}), \bibinfo{pages}{109544--109554}.
\newblock
\showISSN{2169-3536}
\urldef\tempurl%
\url{https://doi.org/10.1109/ACCESS.2019.2933492}
\showDOI{\tempurl}
\newblock
\shownote{Conference Name: IEEE Access}.


\bibitem[Sutton and Barto(2018)]%
        {sutton_reinforcement_2018}
\bibfield{author}{\bibinfo{person}{Richard~S. Sutton} {and} \bibinfo{person}{Andrew~G. Barto}.} \bibinfo{year}{2018}\natexlab{}.
\newblock \bibinfo{booktitle}{\emph{Reinforcement {Learning}: {An} {Introduction}}}.
\newblock \bibinfo{publisher}{A Bradford Book}, \bibinfo{address}{Cambridge, MA, USA}.
\newblock
\showISBNx{978-0-262-03924-6}


\bibitem[Sutton et~al\mbox{.}(1999)]%
        {sutton_policy_1999}
\bibfield{author}{\bibinfo{person}{Richard~S. Sutton}, \bibinfo{person}{David McAllester}, \bibinfo{person}{Satinder Singh}, {and} \bibinfo{person}{Yishay Mansour}.} \bibinfo{year}{1999}\natexlab{}.
\newblock \showarticletitle{Policy gradient methods for reinforcement learning with function approximation}. In \bibinfo{booktitle}{\emph{Proceedings of the 12th {International} {Conference} on {Neural} {Information} {Processing} {Systems}}} \emph{(\bibinfo{series}{{NIPS}'99})}. \bibinfo{publisher}{MIT Press}, \bibinfo{address}{Cambridge, MA, USA}, \bibinfo{pages}{1057--1063}.
\newblock


\bibitem[Toll and Pettré(2021)]%
        {toll_algorithms_2021}
\bibfield{author}{\bibinfo{person}{W. Toll} {and} \bibinfo{person}{J. Pettré}.} \bibinfo{year}{2021}\natexlab{}.
\newblock \showarticletitle{Algorithms for {Microscopic} {Crowd} {Simulation}: {Advancements} in the 2010s}.
\newblock \bibinfo{journal}{\emph{Computer Graphics Forum}} \bibinfo{volume}{40}, \bibinfo{number}{2} (\bibinfo{date}{May} \bibinfo{year}{2021}), \bibinfo{pages}{731--754}.
\newblock
\showISSN{0167-7055, 1467-8659}
\urldef\tempurl%
\url{https://doi.org/10.1111/cgf.142664}
\showDOI{\tempurl}


\bibitem[Whittle(2008)]%
        {whittle_gait_2008}
\bibfield{author}{\bibinfo{person}{Michael~W. Whittle}.} \bibinfo{year}{2008}\natexlab{}.
\newblock \bibinfo{booktitle}{\emph{Gait analysis: an introduction} (\bibinfo{edition}{4th ed., reprinted} ed.)}.
\newblock \bibinfo{publisher}{Butterworth-Heinemann, Elsevier}, \bibinfo{address}{Edinburgh}.
\newblock
\showISBNx{978-0-7506-8883-3}


\bibitem[Xu et~al\mbox{.}(2020)]%
        {xu_local_2020}
\bibfield{author}{\bibinfo{person}{Dong Xu}, \bibinfo{person}{Xiao Huang}, \bibinfo{person}{Zhenlong Li}, {and} \bibinfo{person}{Xiang Li}.} \bibinfo{year}{2020}\natexlab{}.
\newblock \showarticletitle{Local motion simulation using deep reinforcement learning}.
\newblock \bibinfo{journal}{\emph{Transactions in GIS}} \bibinfo{volume}{24}, \bibinfo{number}{3} (\bibinfo{year}{2020}), \bibinfo{pages}{756--779}.
\newblock
\showISSN{1467-9671}
\urldef\tempurl%
\url{https://doi.org/10.1111/tgis.12620}
\showDOI{\tempurl}
\newblock
\shownote{\_eprint: https://onlinelibrary.wiley.com/doi/pdf/10.1111/tgis.12620}.


\bibitem[Xu and Karamouzas(2021)]%
        {xu_human-inspired_2021}
\bibfield{author}{\bibinfo{person}{Pei Xu} {and} \bibinfo{person}{Ioannis Karamouzas}.} \bibinfo{year}{2021}\natexlab{}.
\newblock \showarticletitle{Human-{Inspired} {Multi}-{Agent} {Navigation} using {Knowledge} {Distillation}}.
\newblock \bibinfo{journal}{\emph{arXiv:2103.10000 [cs]}} (\bibinfo{date}{March} \bibinfo{year}{2021}).
\newblock
\urldef\tempurl%
\url{http://arxiv.org/abs/2103.10000}
\showURL{%
\tempurl}
\newblock
\shownote{arXiv: 2103.10000}.


\bibitem[Zheng and Liu(2019)]%
        {zheng_improved_2019}
\bibfield{author}{\bibinfo{person}{S. Zheng} {and} \bibinfo{person}{H. Liu}.} \bibinfo{year}{2019}\natexlab{}.
\newblock \showarticletitle{Improved {Multi}-{Agent} {Deep} {Deterministic} {Policy} {Gradient} for {Path} {Planning}-{Based} {Crowd} {Simulation}}.
\newblock \bibinfo{journal}{\emph{IEEE Access}}  \bibinfo{volume}{7} (\bibinfo{year}{2019}), \bibinfo{pages}{147755--147770}.
\newblock
\showISSN{2169-3536}
\urldef\tempurl%
\url{https://doi.org/10.1109/ACCESS.2019.2946659}
\showDOI{\tempurl}
\newblock
\shownote{Conference Name: IEEE Access}.


\bibitem[Zipf(1949)]%
        {zipf_human_1949}
\bibfield{author}{\bibinfo{person}{George~K. Zipf}.} \bibinfo{year}{1949}\natexlab{}.
\newblock \bibinfo{booktitle}{\emph{Human {Behaviour} and the {Principle} of {Least} {Effort}}}.
\newblock \bibinfo{publisher}{Addison-Wesley}.
\newblock


\end{thebibliography}

\newpage
\appendix
\onecolumn
\section{Discounting-invariant reward}

Intermediate equations in \ref{sec:disc-invariance}. Derivatives and solutions computed using sympy.

Here we provide the intermediate stages of the computation in~\ref{sec:disc-invariance}. All computations are performed using sympy -- finding the derivatives and roots of equations.

First, we reiterate the discounted sum of rewards with a discount factor $\gamma$ and a guiding potential with the coefficient $c_p \sqrt{e_s e_w}$. (Equation~\ref{eq:discounted-reward-sum})

\begin{equation*}
    R^\gamma = \int_0^T e^{t \ln \gamma} \left(-e_s - e_w v^2 + c_p \sqrt{e_s e_w} v\right) dt \nonumber = \frac{1 - \gamma^\frac{d}{v}}{-\ln \gamma} \left( -e_w v^2 + c_p \sqrt{e_s e_w} - e_s\right)
\end{equation*}

We differentiate the expression on the right hand side w.r.t. $v$, looking for a stationary point that corresponds to the optimal velocity $v^*$:

\[
\frac{dR^\gamma}{dv} = F(v, \gamma) = - d \gamma^{\frac{d}{v}} \left(\frac{c_{p} \sqrt{e_{s} e_{w}}}{v} - \frac{e_{s}}{v^{2}} - e_{w}\right) - \frac{\left(1 - \gamma^{\frac{d}{v}}\right) \left(c_{p} \sqrt{e_{s} e_{w}} - 2 e_{w} v\right)}{\log{\left(\gamma \right)}} = 0
\]

While we cannot solve this analytically for $v^*$, we know that arguments for which $F(v^*, \gamma) = 0$ correspond to optimal velocities with a given $\gamma$. We are now interested in finding arguments where $\frac{dv^*}{d\gamma} = 0$, \ie changes in the discount factor do not affect the optimal velocity. We find that by computing the implicit derivative:

\[
\frac{dv}{d\gamma} = \frac{v \left(- d^{2} \gamma^{\frac{d + 2 v}{v}} \left(- c_{p} v \sqrt{e_{s} e_{w}} + e_{s} + e_{w} v^{2}\right) \log{\left(\gamma \right)}^{2} - d v^{2} \gamma^{\frac{d + 2 v}{v}} \left(c_{p} \sqrt{e_{s} e_{w}} - 2 e_{w} v\right) \log{\left(\gamma \right)} + v^{3} \gamma^{2} \left(\gamma^{\frac{d}{v}} - 1\right) \left(c_{p} \sqrt{e_{s} e_{w}} - 2 e_{w} v\right)\right)}{\gamma^{3} \left(d v^{2} \gamma^{\frac{d}{v}} \left(c_{p} \sqrt{e_{s} e_{w}} - 2 e_{w} v\right) \log{\left(\gamma \right)} + d \gamma^{\frac{d}{v}} \left(d \left(- c_{p} v \sqrt{e_{s} e_{w}} + e_{s} + e_{w} v^{2}\right) \log{\left(\gamma \right)} - v \left(c_{p} v \sqrt{e_{s} e_{w}} - 2 e_{s}\right)\right) \log{\left(\gamma \right)} + 2 e_{w} v^{4} \left(\gamma^{\frac{d}{v}} - 1\right)\right) \log{\left(\gamma \right)}} = 0
\]   

Although this expression is even more complex, it turns out to be solvable analytically for $c_p$:

\[
c_p = \frac{d^{2} e_{s} \gamma^{\frac{d}{v}} \log{\left(\gamma \right)}^{2} + d^{2} e_{w} v^{2} \gamma^{\frac{d}{v}} \log{\left(\gamma \right)}^{2} - 2 d e_{w} v^{3} \gamma^{\frac{d}{v}} \log{\left(\gamma \right)} + 2 e_{w} v^{4} \gamma^{\frac{d}{v}} - 2 e_{w} v^{4}}{v \sqrt{e_{s} e_{w}} \left(d^{2} \gamma^{\frac{d}{v}} \log{\left(\gamma \right)}^{2} - d v \gamma^{\frac{d}{v}} \log{\left(\gamma \right)} + v^{2} \gamma^{\frac{d}{v}} - v^{2}\right)} = 2
\]

After substituting numerical values and $v = v^* = \sqrt{\frac{e_s}{e_w}}$, we get the result $c_p = 2$. This means that if the potential has a coefficient of $c_p = 2$, the optimal velocity will not change with the discount factor.



\section{Algorithmic details}

The performance of agents trained with PPO tends to significantly depend on the exact implementation details of the algorithm~\citep{engstrom_implementation_2020, andrychowicz_what_2020, huang_37_2022}. Beyond a set of typical choices used in our implementation (all of which can be found in the training code), we use a non-standard modification that we call ``rewind'', inspired by TRPO.

When performing gradient updates with a given batch of data, the algorithm typically keeps changing the policy until a predefined number of updates elapses. Alternatively, if the KL divergence between the behavior policy and the learned policy exceeds a predefined threshold, the process is stopped immediately to obtain a fresh batch of data.

While this approach typically works sufficiently well for maintaining the on-policy assumption of the policy gradient theorem, sometimes a single gradient update leads to a significant drop in the performance, which would then take many training iterations to recover. To counteract this effect, we save the policy parameters before each gradient update. If the KL divergence criterion is triggered, the policy is rolled back to that saved state, ensuring that a single batch of data never leads to an excessive change to the policy.

\section{Reward implementation details}

Due to various differences in the basic simulation setup, including but not limited to the design choices described by~\citet{kwiatkowski_understanding_2023}, we were unable to fully reproduce some of the results from prior work. Here, we describe the differences between the reward functions described in other papers, and our implementations.

\citet{lee_crowd_2018} use the function they named FLOOD, which linearly penalizes velocities above $\SI{1.5}{\meter/\second}$ and below $\SI{-0.5}{\meter/\second}$. Because our simulation does not allow backwards movement, this is reduced to a linear penalty to velocities exceeding the optimal velocity (which varies by agent).

Work by \citet{xu_human-inspired_2021} focuses on using knowledge distillation for more human-like behavior, but a key component of their reward function deals with the agents maintaining the right speed. The expression listed in the paper is $w_v \exp(\sigma_v ||\vv - \vv^*||)$, with $w_v = 0.08$ and $\sigma_v = 0.85$. Notice, however, that this structure would incentivize large deviations from the optimal velocity by maximizing $||\vv - \vv^*||$. Due to the monotonicity of the exponential function, exactly one of these parameters must be negative to optimize the behavior in the correct dimension. The source code provided with the paper indicates that $w_v = 0.02$ and $\sigma_v = -0.85$, but in our experiments the reverse convention achieves significantly better results, \ie $w_v < 0$ and $\sigma_v > 0$. Furthermore, in our experiments we use $w_v = -10$ together with adjusted goal and collision rewards, because values closer to the original ones failed to converge to reliable goal-seeking behavior.

\end{document}